\documentclass[5p]{elsarticle}
\usepackage[numbers]{natbib}
\usepackage{booktabs}
\usepackage{subfigure}
\usepackage{amsmath}
\usepackage{amssymb}
\usepackage{array}
\usepackage{algorithm}
\usepackage{algorithmic}
\usepackage{color}
\usepackage{caption}

\journal{Robotics and Autonomous Systems}

\begin{document}

\begin{frontmatter}


\title{Real-time Identification and Simultaneous Avoidance of Static and Dynamic Obstacles on Point Cloud for UAVs Navigation\tnoteref{t1}}                 
\tnotetext[t1]{This project is supported by the seed funding for strategic interdisciplinary research scheme of the University of Hong Kong}

\author[1] {Han Chen} 
\ead{stark.chen@connect.polyu.hk}

\address[1]{Department of Aeronautical and Aviation Engineering, Hong Kong Polytechnic University, Hong Kong, China.}

\author[2] {Peng Lu\corref{cor1}}
\ead{lupeng@hku.hk}

\address[2]{Department of Mechanical Engineering, The University of Hong Kong, Hong Kong, China.}

\cortext[cor1]{Corresponding author}

\begin{abstract}
Avoiding hybrid obstacles in unknown scenarios with an efficient flight strategy is a key challenge for unmanned aerial vehicle applications. In this paper, we introduce a more robust technique to distinguish and track dynamic obstacles from static ones with only point cloud input. Then, to achieve dynamic avoidance, we propose the forbidden pyramids method to solve the desired vehicle velocity with an efficient 
sampling-based method in iteration. The motion primitives are generated by solving a nonlinear optimization problem with the constraint of desired velocity and the waypoint.
Furthermore, we present several techniques to deal with the position estimation error for close objects, the error for deformable objects, and the time gap between different submodules. The proposed approach is implemented to run onboard in real-time and validated extensively in simulation and hardware tests, demonstrating our superiority in tracking robustness, energy cost, and calculating time.
\end{abstract}




\end{frontmatter}


\section{Introduction}

In unknown and chaotic environments, unmanned aerial vehicles (UAVs), especially quadcopters always face rapid unexpected changes, while moving obstacles pose a greater threat than static ones. To tackle this challenge, the trajectory planner for UAVs needs to constantly and quickly generate collision-free and feasible trajectories in different scenarios, and its response time is required to be as short as possible. In addition, the optimality of the motion strategies should also be considered to save the limited energy of quadrotors.

\begin{figure}[t]
      \centering
      \includegraphics[width=0.99\linewidth]{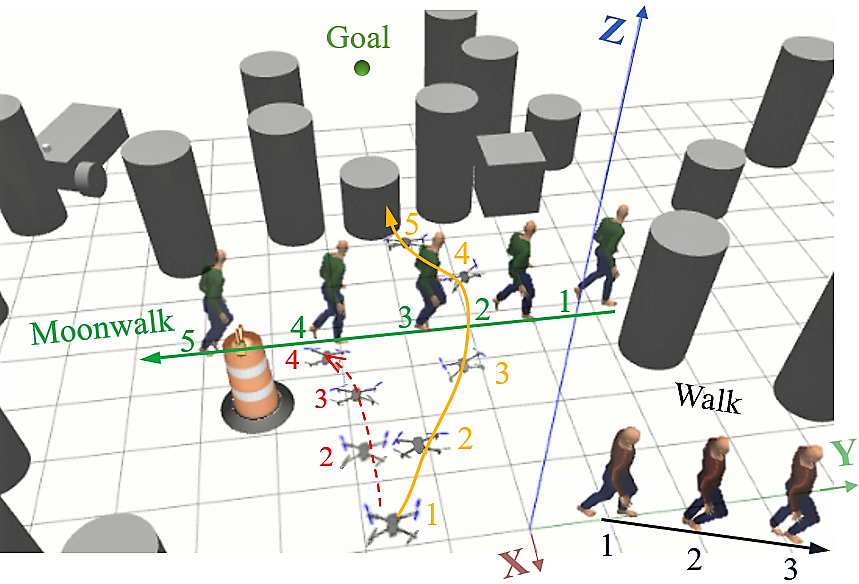}
      \caption{The composite picture of the simulation in Gazebo for the process that the drone avoids static and dynamic obstacles. 5 screenshots are used for composition and the cut time interval is fixed to 0.7 seconds. The line with an arrowhead shows the moving direction and the numbers mark the corresponding frame, the numbers increase by time. The yellow line is generated by the method in this paper, while the red line is by the static planning method.}
      \label{fig1}
      \vspace{-0.05cm}
   \end{figure}

 Most existing frameworks that enable drones to generate collision-free trajectories in completely unknown environments only take into consideration stationary obstacles. However, as quadrotors often fly at low altitudes, they are faced with various moving obstacles such as vehicles and pedestrians on the ground. One primary solution to avoid collision is to raise flight altitude to fly above all the obstacles. This method is not feasible for some indoor applications, because the flight altitude is limited in narrow indoor space, and the drones are often requested to interact with humans as well. Another solution is to assume all detected obstacles as static. But this method cannot guarantee the safety of the trajectory \cite{chen2016online}, considering measurement errors from the sensors and unmissable displacement of the dynamic obstacles. As shown in Figure \ref{fig1}, the collision may happen if the motion information of obstacles is ignored (the red line). Therefore, a more efficient and safer way to avoid moving obstacles is to predict and consider the obstacles’ position in advance based on the velocity, which can avoid detours or deadlock on some occasions.
	
 
To fly in dynamic scenes, we propose a complete system in this paper, composed of a position and velocity estimator for moving obstacles, an upper-level planner to obtain desired velocity, and a motion planner to generate final motion primitives. For the perception of dynamic obstacles, the dynamic ones are identified from static ones by clustering and comparing the displacement from two point cloud data frames. A RGB-D camera is the only sensor utilized to obtain the point cloud. First, we set up a Kalman filter for each dynamic obstacle for tracking and output more accurate and continuous estimating results. The feature vector for each obstacle is adopted to improve the obstacle matching accuracy and robustness, thus the dynamic obstacles tracking and position and velocity estimating performance are improved comparing to the related existing works. In addition, we introduce the track point to reduce the displacement estimation error involved by the self-occlusion of the obstacle.

Then, with the estimated position and velocity of obstacles and the current states and kinodynamic limitations of a real vehicle, the forbidden pyramids method is applied to plan the desired velocity to avoid obstacles. The desired velocity is obtained from a sampling-based method in the feasible space, and the sampled velocity with the minimal acceleration cost is chosen. Finally, the motion primitives are efficiently solved from a well-designed non-linear optimization problem, where the desired velocity is the constraint. For navigation tasks, the proposed velocity planning method is also flexible to combine with most path planning algorithms for static environments, giving them the ability to avoid dynamic obstacles. The waypoint in the path acts as the trajectory endpoint constraints at a further time horizon. In this paper, we test it with our former proposed waypoint planning method heuristic angular search (HAS) \cite{chen2020computationally} to complete the system and conduct the flight tests. The safety and the lower acceleration cost of this method can be verified by the data from flight tests. The computational efficiency of the whole system also shows great advantages over state-of-the-art (SOTA) works \cite{lin2020robust}-\cite{wang2021autonomous}. 

In summary, the main contributions of the paper are as follows:

\begin{itemize}

\item The feature vector is introduced to help match the corresponding obstacle in two point cloud frames. It is proved to be more robust than existing works that match the obstacles with only position information predicted by the Kalman filter. The neighbor frame overlapping and ego-motion compensation techniques are further introduced to reduce the estimating errors of the obstacle's position.
\item{To compensate for the resultant displacement estimation error from the self-occlusion of obstacles, the object track point is proposed.}
\item Based on the relative velocity, the forbidden pyramids method is designed to efficiently plan the safe desired velocity to avoid both dynamic and static obstacles. The varies time gaps which may cause control lag error are also well compensated.
\item We integrate those proposed methods and a path planning method into a complete quadrotor system, demonstrating its reliable performance in flight tests.
\end{itemize}
\section{Related work}

At present, there are many methods for path planning and obstacle avoidance in static environments. Although some researchers have published their safe planning framework for UAVs in an unknown static scenario, it is a more complex problem for a UAV with a single depth camera flying in an unknown environment with dynamic obstacles.

For identifying and tracking moving obstacles from the environment, most researchers employ the raw image from the camera and mark the corresponding pixels before measuring the depth. The semantic segmentation network with a moving consistency check method based on images can distinguish the dynamic objects \cite{yu2018ds}. Also, a block-based motion estimation method to identify the moving obstacle is used in \cite{kim2012moving}, but the result is poor if the background is complex.  Some work \cite{oleynikova2015reactive}-\cite{skulimowski2019interactive} segment the depth image and regard the points with similar depth belong to one object. But such methods cannot present the dynamic environment accurately because static and dynamic obstacles are not classified. If only human is considered as moving obstacle, the human face recognition technology can be applied \cite{nageli2017real}. However, the above-mentioned works do not estimate the obstacle velocity and position.
\cite{ess2009moving} proposes such an approach, which jointly estimates the camera position, stereo depth, and object detections, and track the trajectories. Some works adopt feature-based vision systems to detect dynamic objects \cite{berker2017feature}-\cite{barsan2018robust}, which require dense feature points. Also, detector based or segmentation network based methods can work well in predefined classes such as
pedestrians or cars \cite{zhang2017towards}. However, they cannot handle generic environments. Considering the limited resource of the onboard microcomputer, the above image-based methods are computation-expensive and thus not able to run in real-time. 

Based on the point cloud data, it is also possible to estimate the moving obstacles' position and velocity in the self-driving cars \cite{kraemer2018lidar}-\cite{miller2016dynamic}. However, they all rely on high-quality point clouds from LiDAR sensors and powerful GPUs to detect obstacles from only predefined classes. To enhance the versatility, \cite{cherubini2014autonomous} offers an idea to track point clusters with the global feature in simple environments. For the point cloud of a depth camera, the existing works are rare and they all match the obstacle by only the center position of obstacles, depending on the Kalman filter to predict the position of dynamic obstacles from past to present. However, this may fail when the predicted position of one obstacle is close to other obstacles at present. Varying from them, we propose the feature vector for each obstacle to tackle this challenge, and the matching robustness and accuracy are improved a lot. Our method also can be run at a higher frequency with low computational power. Event cameras can distinguish between static and dynamic objects and enable the drone to avoid the dynamic ones in a very short time \cite{falanga2020dynamic}. However, the event camera is expensive for low-cost UAVs and not compatible to sense the static obstacles.

In terms of the avoidance of moving obstacles for navigation tasks, the majority of research works are based on the applications of ground vehicles. The forbidden velocity map \cite{damas2009avoiding} is designed to solve out all the forbidden 2D velocity vectors and they are represented as two separate areas in the map. The artificial potential field (APF) method can avoid the moving obstacles by considering their moving directions \cite{malone2017hybrid}-\cite{febbo2017moving}. For UAVs, the model predictive control (MPC) method is tested, but the time cost is too large for real-time flight \cite{lin2020robust}. \cite{luo2020multi} proposes the probabilistic safety barrier certificates (PrSBC) to define the space of admissible control actions that are probabilistically safe, which is more compatible for multi-robot systems. Recently, some global planners for UAV navigation in crowded dynamic environment are proposed \cite{cao2019dynamic}-\cite{zhu2020online}. However, the states of all obstacles are known, they are not suitable for a fully autonomous aerial platform. \cite{wang2021autonomous} utilizes the kinodynamic A* algorithm to find a feasible initial trajectory first and the parameterized B-spline is used to optimize the trajectory from the gradient. However, it requires dense samples along the trajectory in the optimization problem, and the object function composes the integration of the whole trajectory. Our motion optimization method is more efficient in computation.

\section{Technical approach}

\begin{figure*}
     \centerline{
      \includegraphics[width=0.9\textwidth]{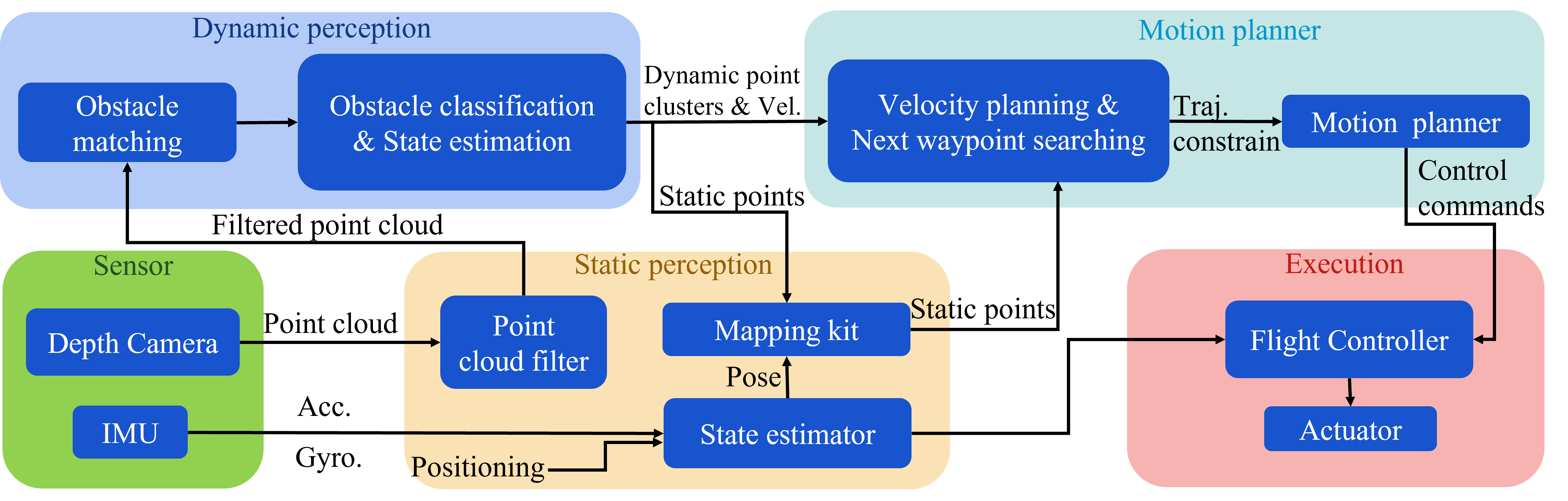}}
      \caption{The proposed system for the autonomous navigation in dynamic environments. The positioning can be done by the outer motion capture system or onboard VIO toolkit.}
      \label{fig_frame}
   \end{figure*}

Our proposed framework is composed of two submodules that run parallelly and asynchronously: the obstacle classifier and motion state estimator (section 3.1   \& 3.2) and the waypoint and motion planner (section 4.1 \& 4.2). The additional technical details for improving the accuracy of dynamic perception are introduced in section 3.3. Figure \ref{fig_frame} illustrates the whole framework, including the important message flowing between the submodules.

\subsection{Obstacle tracking}
The raw point cloud is first filtered to remove the noise and converted into earth coordinate $E\!-\!X\!Y\!Z$. The details about the filter are in section 5. We use $Pcl_{t_1}$ and $Pcl_{t_2}$ to denote two point cloud frames from the sensor at the former timestamp $t_1$ and the latest timestamp $t_2$ respectively. $t_{2}-t_{1}=d_{t}$ and $d_{t} > 0$. The time interval $d_{t}$ is set to be able to make the displacement of the dynamic obstacles obvious enough to be observed, while maintaining an acceptable delay to output the estimation results. $Pcl_{t_1}$ and $Pcl_{t_2}$ are updated continuously while the sensor is operating. To deal with the movement of the camera between $t_1$ and $t_2$, $Pcl_{t_2}$ is filtered, only keeping the points in the overlapped area of the camera's FOVs at $t_1$ and $t_2$ \cite{wang2021autonomous}. The newly appeared obstacles in the latest frame are removed, so only the obstacles appear in both of the two point cloud frames are further analyzed. $Pcl_{t_1}$ and $Pcl_{t_2}$ are clustered into individual objects using density-based spatial clustering of applications with noise (DBSCAN) \cite{ester1996a}, resulting in two sets of clusters $OB_{1}=\{ob_{11},ob_{12},...\}$ and $OB_{2}=\{ob_{21},ob_{22},...\}$. Then, matching the two clusters $ob_{2k} \in OB_{2}$ and $ob_{1j} \in OB_{1}$ corresponding to the same obstacle is necessary before the dynamic obstacle identification. 

\begin{figure}[t]
      \centering
      \includegraphics[width=0.99\linewidth]{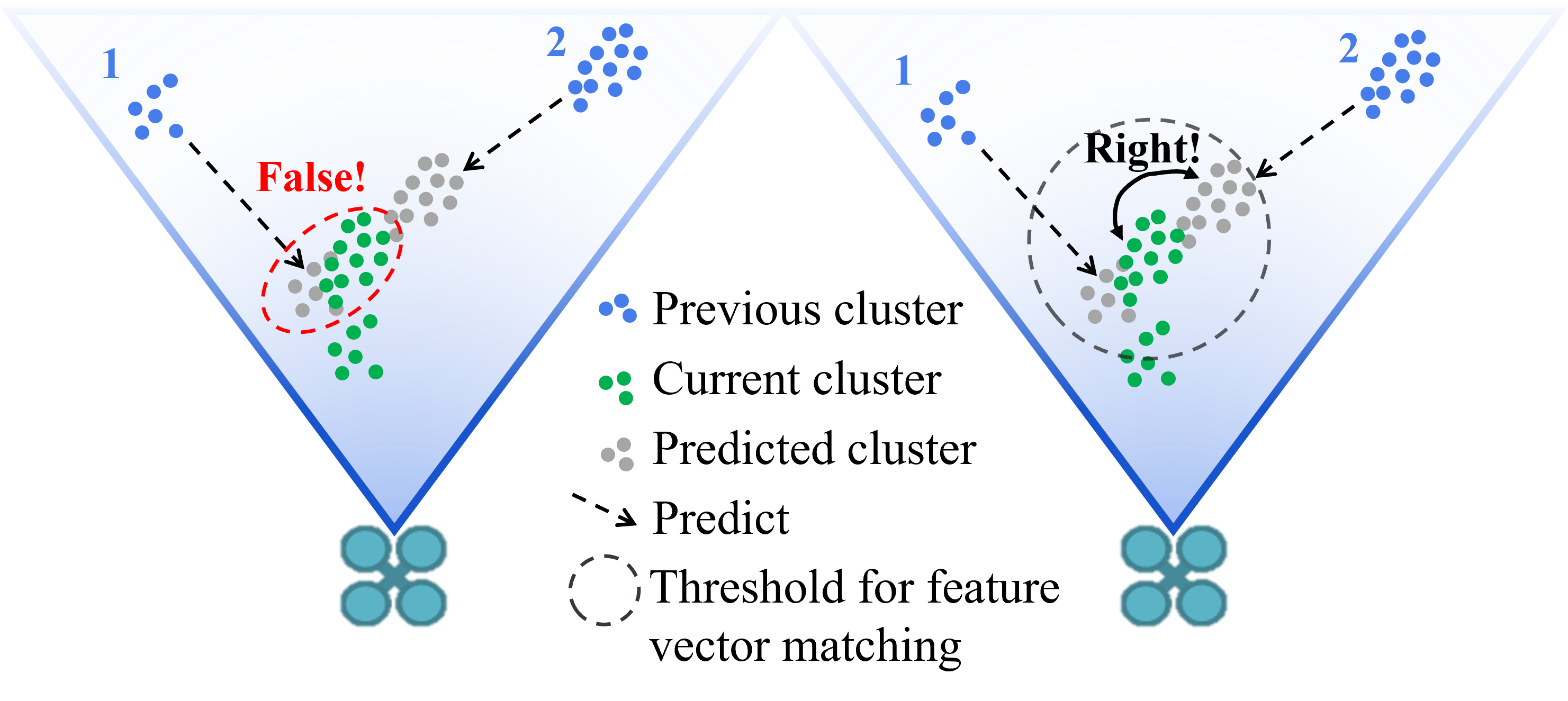}
      \caption{The left figure shows a situation that two obstacles are mismatched. The predicted cluster of obstacle 1 is closer than the predicted cluster 2 to the current cluster of obstacle 2. By comparing the feature vector, the correct predicted cluster for obstacle 2 can be matched for the current cluster, as shown in the right figure.}
      \label{fig_featurevec}
      \vspace{-0.0cm}
   \end{figure}
   
At the time when we obtain the first frame of $Pcl_{t_2}$, a list of Kalman filters with the constant velocity model is initialized for each cluster (obstacle) in $OB_{2}$. The position and velocity are updated after the obstacle is matched and the observation values are obtained. To match the obstacle, we preliminarily sort out the two clusters that satisfies the condition $\|pos(ob_{2k})-pos(ob_{1j})\|_{2} < d_{m}$. $d_m$ is the distance threshold. $pos() \in \mathbb{R}^3$ gives the obstacle position when the input cluster is from the latest frame $Pcl_{t_2}$, while it returns the predicted position at $t_2$ by the corresponding Kalman filter of the cluster from $Pcl_{t_1}$ \cite{wang2021autonomous}. It is to associate current clusters to the forward propagated Kalman filters rather than clusters in the previous frame. If we cannot find an $ob_{1j}$ for $ob_{2k}$ there,
we skip it and turn to the next cluster ($k \leftarrow k+1$, $k$ is the cluster index). For each Kalman filter, it is also necessary to assign a reasonable maximal propagating time $t_{kf}$ before being matched with a new observation. Because the camera FOV is narrow, we hope to predict the clusters which just move out of the FOV for safety consideration, and they are assumed to continue to move at their latest updated velocity in a short period. A Kalman filter is deleted together with its tracking history if it has not been matched for over $t_{kf}$.

However, matching obstacles with position only may fail when obstacles are getting close, as shown in Figure \ref{fig_featurevec}. To improve the matching robustness, we design the feature vector composed of several statistic characters of a point cluster with aligned color information from the obstacle, which is defined as
$$ fte(ob) = [len(ob) ,V^{A}_{P}(ob), V(ob), M_{C}(ob),V^{A}_{C}(ob)], \eqno{(1)}$$
where $ob$ denotes any point cluster. $len()$ is the function that returns the size of the input cluster. $V^{A}_{P}() \in \mathbb{R}^3$ returns the position variance of the cluster, and $V()$ returns the volume of the axis-aligned bounding box (AABB) of the cluster. $M_{C}() \in \mathbb{R}^3$ and $V^{A}_{C}() \in \mathbb{R}^3$ return the mean and variance of the RGB value of points. The idea is: if there is not a significant difference in the shape and color of the two point clusters extracted from two timely close point cloud frames respectively, then they are commonly believed to be the same object. The global features for each obstacle are very cheap to calculate and proved to be effective in tests.

At last, the Euclidean distance $d_{fte}=\|fte(ob_1j)-fte(ob_{2k})\|_{2}$ between feature vectors is calculated with each cluster pair $\{ ob_{2k}, ob_{1j}\}$ that satisfies the position threshold. For each cluster $ob_{2k} \in OB_{2}$, the cluster $ob_{1J} \in OB_{1}$ results in the minimal $d_{fte}$ is matched with it. The feature vector is normalized to 0-1 since the order of magnitude of each element varies. We use $ob^{m}_{2} \in OB_{2}$ and $ob^{m}_{1} \in OB_{1}$ to represent any two successfully matched clusters.

\subsection{Obstacle Classification}
\begin{figure}[t]
      \centering
      \includegraphics[width=0.99\linewidth]{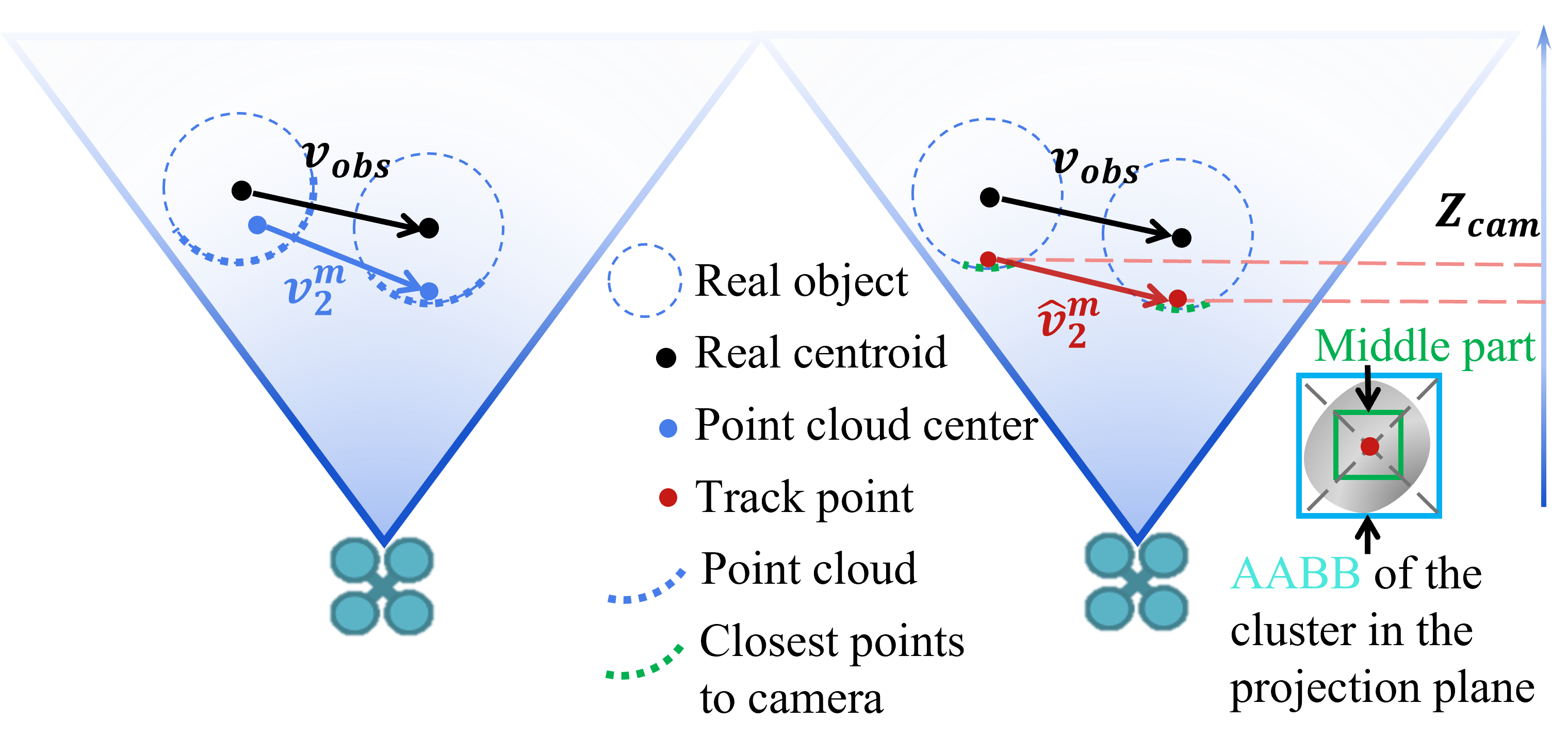}
      \caption{The left figure illustrates the velocity estimation error caused by the self-occlusion of obstacle. $v_{obs}$ is the velocity ground truth. When the obstacle approaches the camera, the visible part shrink, resulting in the relative displacement between the point cloud center and obstacle centroid. The track point in the right figure can reduce the velocity estimation error. The middle part of the cluster is bounded by the green box.}
      \label{fig_trackpoint}
      \vspace{-0.0cm}
   \end{figure}

After the obstacles in two sensor frames are matched in pairs, the velocity of the obstacle $ob^{m}_{2}$ can be calculated by $ v^{m}_{2} = \overrightarrow {p^{m}_{2}p^{m}_{1}}/{(t_{2}-t_{1})}$, where $p^{m}_{2}$ and $p^{m}_{1}$ are the position of the corresponding obstacle. In the related works, the mean of the points in each cluster is adopted as the obstacle position, since most of the obstacles in the UAVs application scenarios can be regarded as mass-uniformly distributed. However, due to the self-occlusion, the backside of the obstacle is invisible, so the mean of points is closer to the camera than the obstacle centroid. In addition, the occluded part of a moving obstacle changes when the relative movement occurs between the camera and obstacle. Thus, the relative position between the position mean and the real mass center also changes. This will lead to a wrongly estimated velocity, as shown in Figure \ref{fig_trackpoint}, the error is mainly distributed on the $Z$ axis of the camera $Z_{cam}$ and it is not a fixed error can be estimated. For the position estimation of obstacles, the self-occlusion is not important, considering the visible part only is safe for obstacle avoidance. But the velocity is the key information of moving obstacles in the vehicle velocity planning. Therefore, we introduce a method to reduce this velocity estimation error by choosing the appropriate track point for the matched pair of clusters, as illustrated in the right figure of Figure \ref{fig_trackpoint}. For a moving obstacle in translational motion, the closest part to the camera is believed not to be self-occluded in $ob^{m}_{2}$ and $ob^{m}_{1}$. In addition, the middle part of the cluster is close to the centroid of the common obstacles, so the rotational movement is weak in this part. Thus, we only use the center point $\hat{p}^{m}_2$ and $\hat{p}^{m}_1$ of the closest $N_{c}$ points in the middle part of the cluster $ob^{m}_{2}$ and $ob^{m}_{1}$ to estimate the displacement. $\hat{p}^{m}_2$ and $\hat{p}^{m}_1$ are named as the track point. Here the distance to the camera is measured only along $Z_{cam}$. The middle part of the cluster is divided in the projection plane corresponding to the depth image. The bounding box for the middle part is shrunk from the AABB of the obstacle to the center proportionally. The shrinking scale factor is $\alpha \ (\alpha < 1)$. Considering the common obstacles usually performs slow rotation, and the time gap $d_t$ is small, we can neglect the influence to the closest part caused by rotation in the displacement estimation. To update the Kalman filter, $p^{m}_{2}$ (the mean of current cluster) is still the observed position, but the velocity observation is $ \hat{v}^{m}_{2} = \overrightarrow {\hat{p}^{m}_{1}\hat{p}^{m}_{2}}/{(t_{2}-t_{1})}$. The classification for static and dynamic obstacle is done by comparing the velocity magnitude with a pre-assigned threshold $v_{dy}$, i.e. $\|\hat{v}^{m}_{2}\|_{2} > v_{dy}$ indicates a moving obstacle. If an obstacle is classified as static in $S_{c}$ consecutive times, the corresponding Kalman filter is abandoned and the static point cluster is forwarded for map fusion.

The Kalman filter for one cluster is detailly described with
$$
\hat{x}_{t}^{-}=F_{t} \hat{x}_{t-1}+B_{t} a_{t-1}, \eqno{(2)} $$ 
$$
P_{t}^{-}=F_{t} P_{t-1} F_{t}^{T}+Q, \eqno{(3)} $$
$$
K_{t}=P_{t}^{-} H^{T}\left(H P_{t}^{-} H^{T}+R\right)^{-1}, \eqno{(4)} $$
$$
P_{t}=\left(I-K_{t} H\right) P_{t}^{-}, \eqno{(5)} $$ 
$$
\hat{x}_{t}=\left\{\begin{array}{l} \hat{x}_{t}^{-}+K_{t}\left(x_{t}-H \hat{x}_{t}^{-}\right) \ \text{(found dynamic obstacle)},\\
\hat{x}_{t}^{-}\  \text{(no dynamic obstacle)}, 
\end{array}\right. \eqno{(6)} $$

$$
x_{t}= \left[\begin{array}{l}
p^{m}_{2} \\
\hat{v}^{m}_{2}
\end{array}\right],F_{t}=\left[\begin{array}{cc}
1 & \Delta t \\
0 & 1
\end{array}\right], B_{t}=\left[\begin{array}{c}
\frac{\Delta t^{2}}{2} \\
\Delta t
\end{array}\right], \eqno{(7)} $$
where $R,\ H,\ Q$ are the observation noise covariance matrix, observation matrix, and error matrix respectively. The superscript $^-$ indicates a matrix is before being updated by the Kalman gain matrix $K_t$, applicable for the state matrix $\hat x_t$ and the posterior error covariance matrix $P_{t}$. The subscripts $t$ and ${t\!-\!1}$ distinguish the current and the former step of the Kalman filter. $F_{t}$ is the state transition matrix and $B_t$ is the control matrix. $x_t$ is composed of the observation of the obstacle position and velocity, and $\hat \ $ marks the filtered results for $x_t$. $\hat{x}_{t}$ equals to the predicted state $\hat{x}_{t}^{-}$ if no dynamic obstacle is caught. $\hat{x}_{t}^{-}$ is also utilized as the propagated cluster state in the obstacle matching introduced above. $\Delta t$ is the time interval between each run of the Kalman filter. In the current stage, we assume the moving obstacle performs uniform motion between $t_1$ and $t_2$, $a_{t-1}=0$.

We summarize our proposed dynamic environment perception method in \textbf{Algorithm \ref{alg2}}.

\begin{algorithm}[!h]
\caption{Dynamic environment perception}
\label{alg2}
\begin{algorithmic}[1]
\WHILE{true:}
\STATE Obtain $Pcl_{t_1}$, $Pcl_{t_2}$ from the point cloud filter, cluster them to $OB_{1}$, $OB_{2}$
\IF {it is the first loop}
\STATE Initialize the Kalman filters for each cluster in $OB_{2}$
\ENDIF
\STATE Predict the position of former clusters with the Kalman filters
\FOR {$ob_{2k}$ in $OB_2$ ($k$ is the iteration number):}
\STATE Match $ob_{2k}$ with the predicted clusters
\IF{successfully match the clusters:}
\STATE Estimate the velocity of $ob_{2k}$ with the paired $ob_{1j}$
\STATE Classify it as static or dynamic and record the class as the history together with the corresponding Kalman filter
\IF{the cluster marked as static for $S_c$ consecutive times:}
\STATE Delete the corresponding Kalman filter, submit $ob_{2k}$ for map fusion
\ELSIF{$ob_{2k}$ is dynamic:}
\STATE Update the Kalman filter with $p^{m}_{2}$ and $\hat v^{m}_{2}$
\ENDIF
\ENDIF
\ENDFOR
\IF{no dynamic obstacle is found:}
\STATE Update the Kalman filter with the predicted state
\ENDIF
\ENDWHILE
\end{algorithmic}
\end{algorithm} 
\vspace{-0.0cm}


\subsection{Ego-motion compensation and neighbor data overlapping}

To improve the estimation accuracy, we notice the time gap between the latest point cloud message from the camera and the vehicle state message from the IMU in the flight controller. A constant acceleration motion model is adopted to describe the vehicle here, and the ego-motion compensation can be done with 
$$\hat{p}_{cam} = p_{cam} + v_{cam}t_{gap} + \frac{1}{2}a_{cam}t^{2}_{gap}, \eqno{(8)}$$
to result in the compensated camera pose $\hat{p}_{cam}$. $p_{cam}  \in \mathbb{R}^{2*3}$, $v_{cam}  \in \mathbb{R}^{2*3}$ and $a_{cam} \in \mathbb{R}^{2*3}$ are the pose, velocity and acceleration of the camera obtained from raw data (translational and rotational motion), translated from the installation matrix of the camera.
$t_{gap}$ is the time gap between the message, equals to the timestamp of point cloud minus the timestamp of vehicle state. As a result, the point cloud can be converted to $E\!-\!X\!Y\!Z$ more precisely.

For non-rigid moving obstacles, for example, walking animals (including humans), the body posture is continuously changing. The point cloud deformation may cause additional position and velocity estimation error of obstacle since the waving limbs of a walking human interfere with the current estimation measurements. We notice that when two neighbor frames of point cloud are overlaid, the point cloud of the human trunk is denser than the other parts which rotate over the trunk. Then, an appropriate point density threshold of DBSCAN can remove the points corresponding to the limbs. So the overlapped point cloud can replace the filtered raw point cloud. For instance, the $w^{th}$ point cloud frame $Pcl_{w}$ is replaced with $Pcl_{w}\cup Pcl_{w-1}$. $\hat{p}_{cam}$ is also replaced by the mean of the value from its neighbor data frame, to align with the point cloud.

\section{Motion planning}

The motion of the drone is more aggressive for avoiding moving obstacles than flying in a static environment. To address the displacement of the drone during the time costed by the trajectory planner and flight controller, position compensation is adopted before the velocity planning. The current position of drone $p_{n}$ is updated by the prediction
\vspace{-0.0cm}
$$\hat {p}_{n} = p_{n} +( t_{pl} + t_{ct} + t_{pm})v_{n} + \frac{1}{2}( t_{pl} + t_{ct} + t_{pm})^2 a_{n},
\eqno{(9)}$$
where $t_{pl}$ is the time cost of the former step of the motion planner. $t_{ct}$ is the fixed time cost for the flight controller. $t_{pm}$ is the timestamp gap between UAV pose and current time.
In addition, due to the time cost of obstacle identification and communication delay, the timestamp on the information of dynamic obstacles is always delayed than that of the pose and velocity message of the drone. Based on the uniform acceleration assumption, the obstacle position $\hat p^{m}_{2}$ in the planner at the current time is predicted and updated as
$$ \hat p^{m}_{2} =  p^{m}_{2} + ( t_{pl} + t_{ct} + t_{pm} + t_{dp} )\hat v^{m}_{2},
\eqno{(10)}$$
where $ t_{dp}$ is the timestamp gap between UAV pose and the received dynamic clusters. The dynamic clusters published by the perception module share the same timestamp with the latest point cloud $pcl_{t2}$.

\subsection{Velocity planning}

\begin{figure}[ht]
      \centering
      \includegraphics[width=0.93\linewidth]{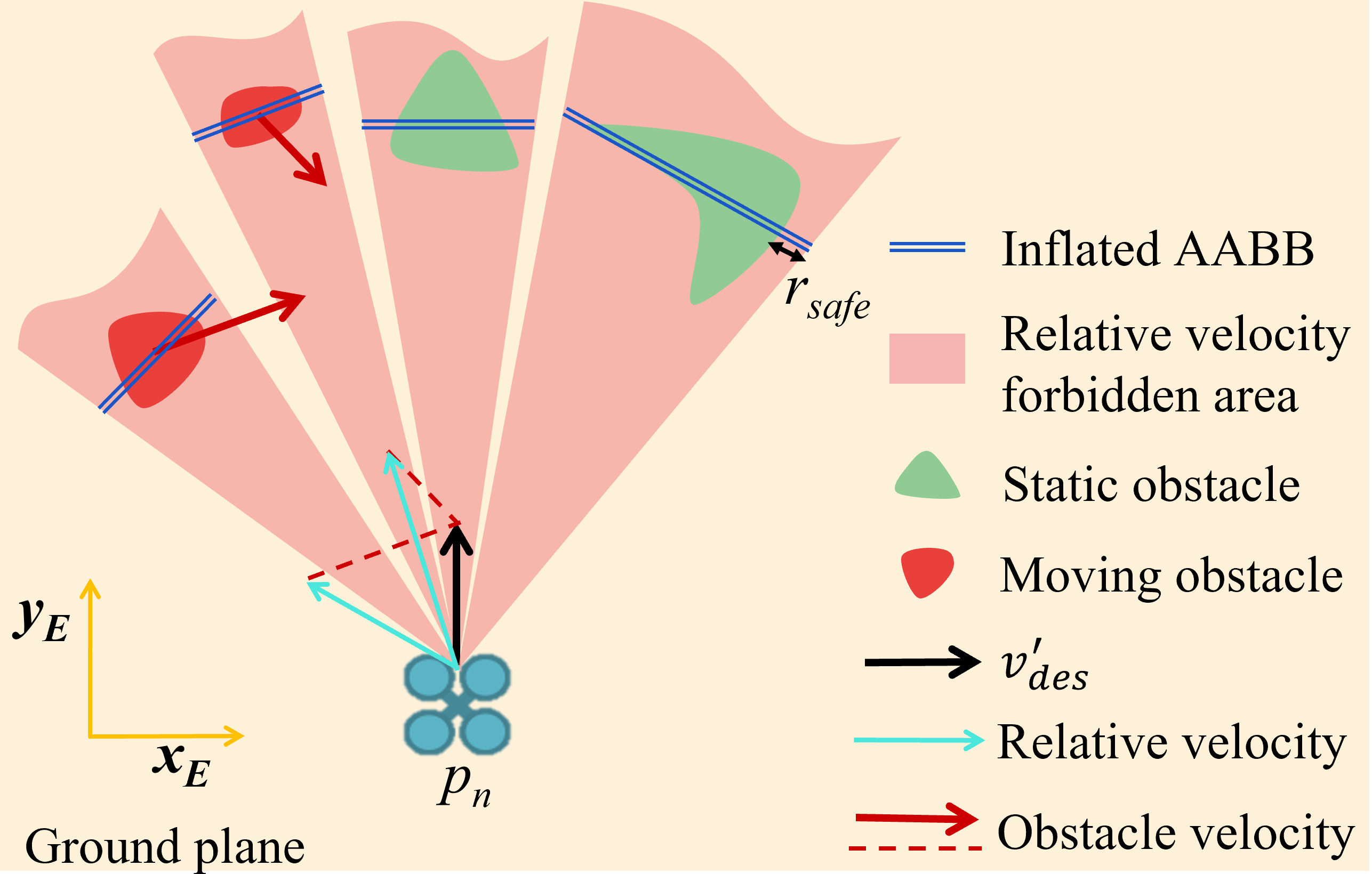}
      \caption{Check if the current relative velocities towards each obstacle lies in the forbidden area. In this figure, the relative velocity towards one dynamic obstacle and one static obstacle all fail the collision check. The forbidden area is the projection area (space) of the inflated obstacle AABB in the projection plane to the camera. $r_{safe}$ is the inflating size.}
      \label{fig_relativevel}
      \vspace{-0.0cm}
   \end{figure}

The planner receives the moving clusters and the velocity from the dynamic perception module. The remaining clusters are also conveyed to the planner and treated as static obstacles. In addition, we adopt the mapping kit to offer the static environmental information out of the current FOV, because the FOV of a single camera is narrow. To tackle the autonomous navigation tasks, the drone is required to reach the goal position. The desired velocity of the drone is initialized as $v'_{des}$, $\|v'_{des}\|_{2} = v_{max}$ and $v'_{des}$ heads towards the goal. $v_{max}$ is the maximum speed constraints. Also, considering the path optimality, a path planner is usually adopted in the navigation. Thus, the waypoint $w_{p}$ generated from the planned path is used to replace the navigation goal if a path planner is required. Otherwise, $w_{p}$ denotes the navigation goal. $w_{p}$ can be generated from a guidance law or assigned directly as the first waypoint in the path to enable the drone follow the path. It is a choice to combine the velocity planning method with the path planning algorithms to adapt to navigation applications better. Figure \ref{fig_relativevel} illustrates the collision check by calculating the relative velocity of $v'_{des}$ towards the obstacles, and if the check fails the velocity re-planning will be conducted. For dynamic obstacles, the relative velocity equals $v'_{des}$ minus the obstacle velocity. For static obstacles, the relative velocity is $v_n$ itself. If $v'_{des}$ is checked to be safe, the finally desired velocity $v_{des}$ is given by $v'_{des}$.

  \begin{figure}[ht]
      \centering
      \includegraphics[width=0.97\linewidth]{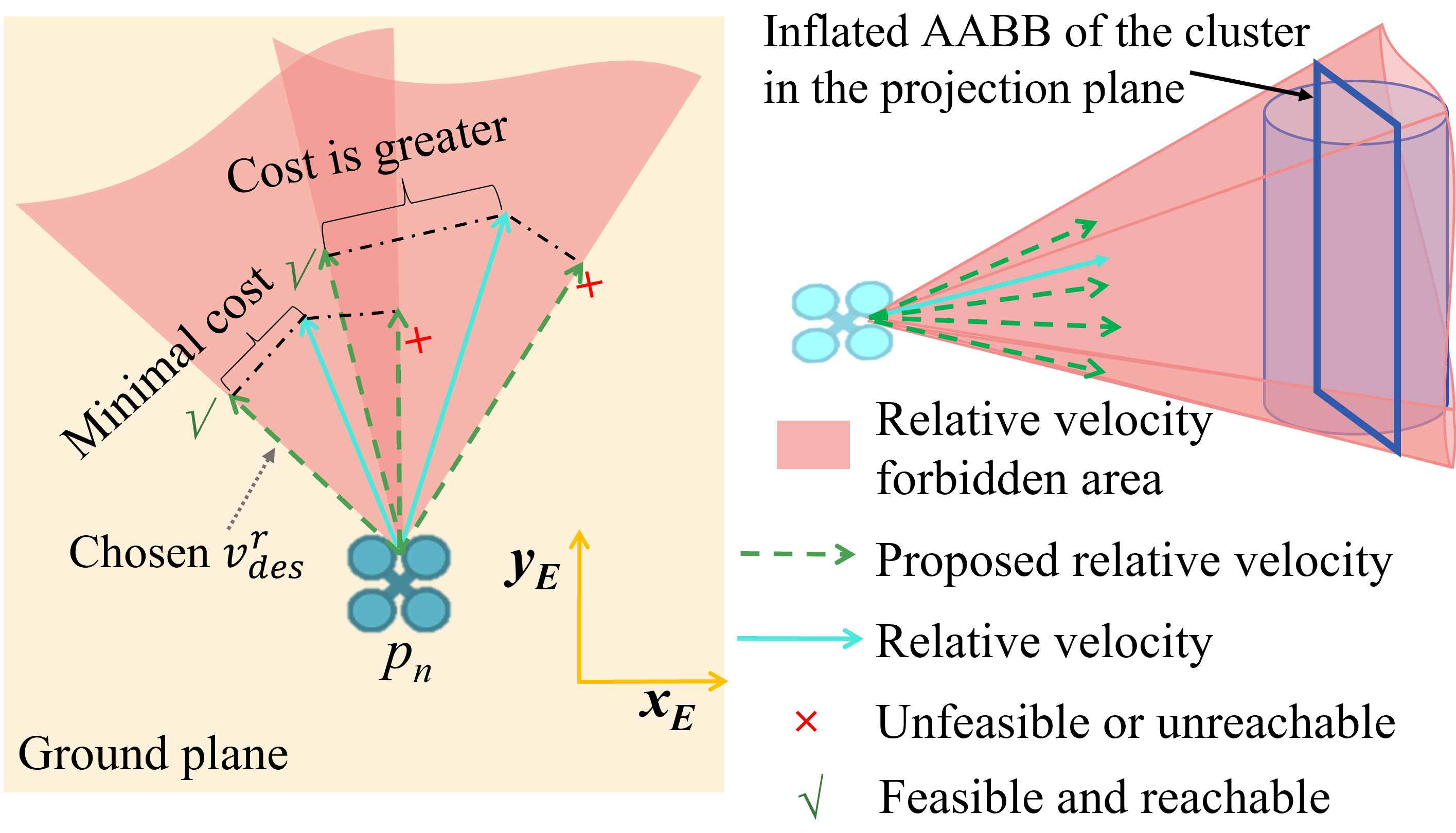}
      \caption{The left figure explains the velocity planning for multiple forbidden pyramids. We use a floor plan to better demonstrate the method. The right figure is a forbidden pyramid for one obstacle in 3D view, four sides of the pyramid result in four proposed relative velocity vectors.}
      \label{fig_velplan}
      \vspace{-0.0cm}
   \end{figure}
 
    \begin{figure}[ht]
      \centering
      \includegraphics[width=0.99\linewidth]{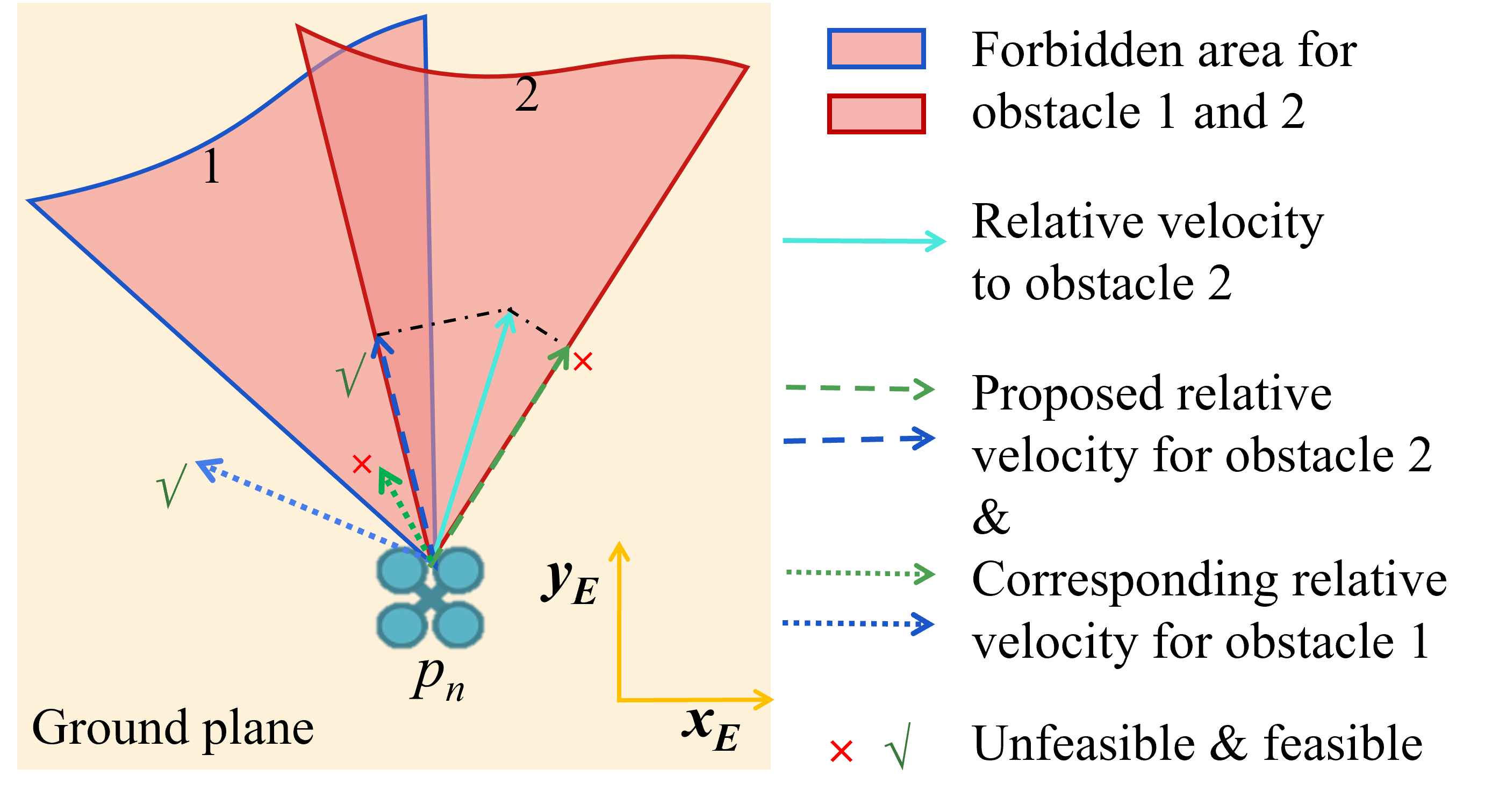}
      \caption{The feasibility check of the relative velocity for one obstacle, as the supplementary for Figure \ref{fig_velplan}. The proposed relative velocity is checked if feasible for other moving obstacles. In this figure, the left proposed relative velocity for obstacle 2 is also feasible for obstacle 1 (navy blue arrows), while the right one (green arrows) is not.}
      \label{fig_velplan1}
      \vspace{-0.3cm}
   \end{figure}
   
     \begin{figure}[ht]
      \centering
      \includegraphics[width=0.40\textwidth]{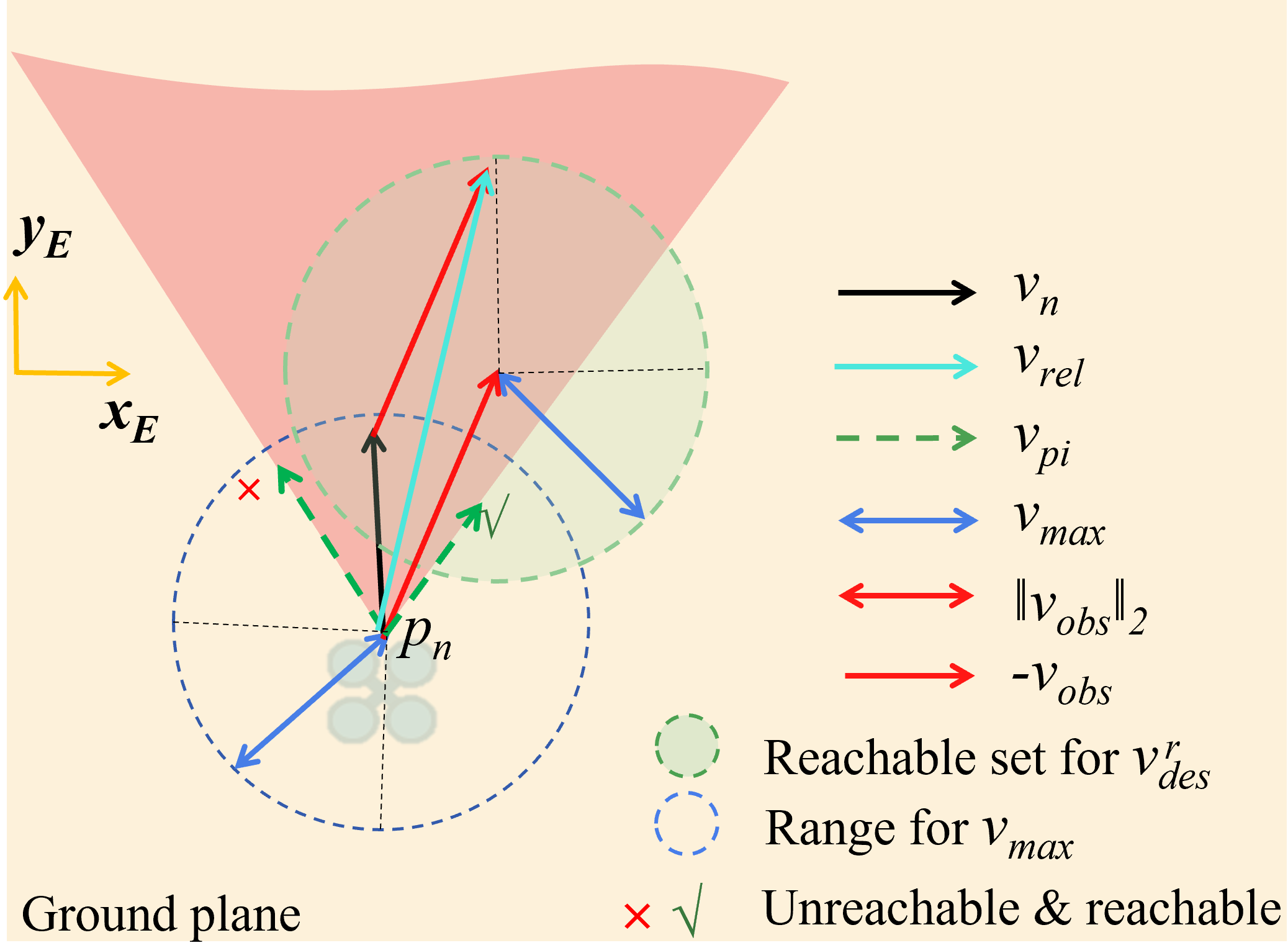}
      \caption{The reachable check for the proposed relative velocity. $v_{obs}$ is moved to start from $p_n$, and the endpoint is the center of the spherical reachable set. All the possible relative velocity constrained by $v_{max}$ towards this obstacle is included in this set. Only the relative velocity vectors in the reachable set are chosen.   
      }
      \label{fig_reachable}
      \vspace{-0.2cm}
   \end{figure}

The velocity re-planning method is explained in Figures \ref{fig_velplan}-\ref{fig_reachable}, the forbidden area (space) is extended to a pyramid for 3D case, different from the triangle for 2D case. If the relative velocity lies in the corresponding forbidden pyramid, four proposed relative velocity vectors are found by drawing a vertical line to the four sides of the pyramid from the relative velocity vector. They are the samples to be checked later. The vertical line segments stand for the acceleration cost to control the vehicle to reach the proposed velocity. For any cluster, $v_{rel}$ denotes the relative velocity, the five vertices of the forbidden pyramid are $$\{\hat{p}_{n}(x_{0},y_{0},z_{0}),vt_{1}(x_{1},y_{1},z_{1}),...,vt_{4}(x_{4},y_{4},z_{4})\}.\eqno{(11)} $$ 
The acceleration cost $a_{ci}$ of the proposed relative velocity vectors $v_{pi}\ (i \in \{1,2,3,4\})$ can be calculated by solving the 3D geometric equations, as follows:
$$\begin{array}{c}
C^{p} =  \left[\begin{array}{c}
(y_{1}-y_{0})(z_{2}-z_{0})-(y_{2}-y_{0})(z_{1}-z_{0}) \\
(z_{1}-z_{0})(x_{2}-x_{0})-(z_{2}-z_{0})(x_{1}-x_{0}) \\
(x_{1}-x_{0})(y_{2}-y_{0})-(x_{2}-x_{0})(y_{1}-y_{0})\end{array}\right]^{T}
\end{array},\eqno{(12)} $$

$$
c^{p}_{d} = -C^{p}{\hat{p}_{n}}^{T},
\eqno{(13)}$$

$$
a_{ci} = \frac{|C^{p}(v_{rel}+\hat{p}_{n})^{T}+c^{p}_{d}|}{\|C^{p}\|_{2}},
\eqno{(14)} $$

$$
v_{pi} = v_{rel} -C^{p}\frac{C^{p}(v_{rel}+\hat{p}_{n})^{T}+c^{p}_{d}}{\|C^{p}\|^{2}_{2}}.
\eqno{(15)} 
$$
In (12)-(15), triangle $\{\hat{p}_{n},vt_{1}, vt_{2}\}$ is taken as the example, $C^{p}[x,y,z]^{T}+c^{p}_{d} = 0$ is the corresponding plane equation, $\hat{p}_{n}$ is the common vertex of all the 4 triangles.

Obviously, for only one obstacle, the desired relative velocity with the minimal cost is from the four proposed ones. For multiple obstacles and forbidden pyramids, the desired relative velocity is chosen by comparing the feasibility, reachability and cost. Among all the proposed relative velocity vectors, the one checked to be feasible and reachable and with the minimal cost is selected ($v^{r}_{des}$), and the desired vehicle velocity $v_{des} = v^{r}_{des} + v_{obs}$. $v_{obs}$ is the velocity for the corresponding obstacle. Although the globally optimal solution for acceleration cost cannot be guaranteed within the samples, the computation complexity is greatly reduced comparing to solve the optimal solution. We use the inflated bounding box because the character radius of the vehicle $r_{uav}$ can not be neglected. 

The feasibility check is to guarantee $v_{des}$ is safe for all obstacles, not for only one of them, which is described in Figure \ref{fig_velplan1}.
Besides the feasibility check, $v_{des}$ should satisfy the maximum speed constraints $v_{max}$. We introduce the reachable set for the relative velocity vector to check if the proposed relative velocity $v_{pi}\ (i \in \{1,2,3,4\})$ is reachable, as Figure \ref{fig_reachable} indicates. For the relative velocity towards one obstacle, $v_{rel} = v_{n} - v_{obs}$ always hold. $v_{rel}$ is the current relative velocity towards the obstacle, $v_{obs}$ is the obstacle velocity.

In addition, the lag error of the velocity planning caused by the time cost to reach the desired velocity is also considerable. The relative displacement between the moving obstacle and vehicle during this time gap should be estimated, because the forbidden pyramid is also directly related with the relative position. We can assume the solved jerk $J_{n}$ is very close to its boundary $J_{max}$, because the time cost $t_{v}$ is minimized in the optimization problem (20). Thus, the time cost is estimated as
$$\begin{aligned}
 & v_{n} + a_{n}t_{v} + \frac{1}{2}J_{n}t_{v}^2 = v_{des} \\
\Rightarrow  t_{v} =  \min(\{t_{v} |\|  &  \frac{2(v_{des}-v_{n}-a_{n}t_{v})}{t_{v}^2}\|_{\infty}=J_{max}\land t_{v}>0\}),\end{aligned}\eqno{(16)}$$
and the displacement of the vehicle and obstacle is calculated by
$$ d_{uav} = v_{n}t_{v} + \frac{1}{2}a_{n}t_{v}^2 + \frac{1}{6}J_{n}t_{v}^3,  \eqno{(17)}$$
$$ d_{obs} = \hat{v}_{2}^{m}t_{v}.   \eqno{(18)}$$

At last, the estimated displacement $d_{uav}$ and $d_{obs}$ are added to the position after $v_{des}$ is obtained, and a new $v_{des}$ is planned in iteration until it is checked to be safe. The accurate time cost $t_{v}$ can only be determined after solving the motion optimization problem. However, involving the optimization problem in the iteration will be time-consuming, so we use a close-form solution as the approximate value. To speed up the convergence, the safety radius is inflated by a small value $\epsilon$ (equivalent to the tolerance in the safety check) to calculate $v_{pi}$: 
$$r_{safe} = r_{uav} + \epsilon\  (\epsilon>0). \eqno{(19)}$$ 

For a situation where the obstacles are too dense, the forbidden area may cover all the space around the vehicle. We first sort all the clusters with the increasing order of distance to $p_n$, the farther obstacles are considered less threatening for the drone. Then, the $j^{th}$ and the next clusters are excluded, $j$ is the iteration number increasing from 0. Algorithm \ref{alg5} reveals the process of velocity planning. As a result, the vehicle can always quickly plan the velocity to avoid static and dynamic obstacles and following the path to meet the different task requirements.
\begin{algorithm}[h]
\caption{Velocity planning} 
\label{alg5}
\begin{algorithmic}[1]
\STATE $v'_{des} \leftarrow v_{max}\dfrac{\overrightarrow{\hat{p}_{n}w_{p}}}{\hat{p}_{n}w_{p}}$
\IF{$v'_{des}$ is unsafe (Figure \ref{fig_relativevel}):}
\STATE $j \leftarrow 0$
\STATE Sort the clusters with the distance to $p_n$
\WHILE{$v_{des}$ is not found:}

\STATE Remove the last $j$ clusters from original sequence 
\STATE Get all the feasible relative velocity vectors for the remained clusters
\IF{feasible and reachable relative velocity exist:}
\STATE Choose $v^{r}_{des}$ with the minimal acceleration cost, and $v_{des} \leftarrow v^{r}_{des} + v_{obs}$
\ENDIF
\STATE $j \leftarrow j+1$
\ENDWHILE
\REPEAT
\STATE $\hat{p}_n \leftarrow \hat{p}_n + d_{uav}$, $\hat{p}_{2}^{m} \leftarrow \hat{p}_{2}^{m}+d_{obs}$
\STATE Repeat line 7-10 with updated forbidden pyramids
\UNTIL{$v_{des}$ is safe}
\ELSE
\STATE $v_{des} \leftarrow v'_{des}$
\ENDIF
\end{algorithmic}
\end{algorithm}  
\subsection{Motion planning}

\begin{figure}[ht]
      \centering
      \includegraphics[width=0.99\linewidth]{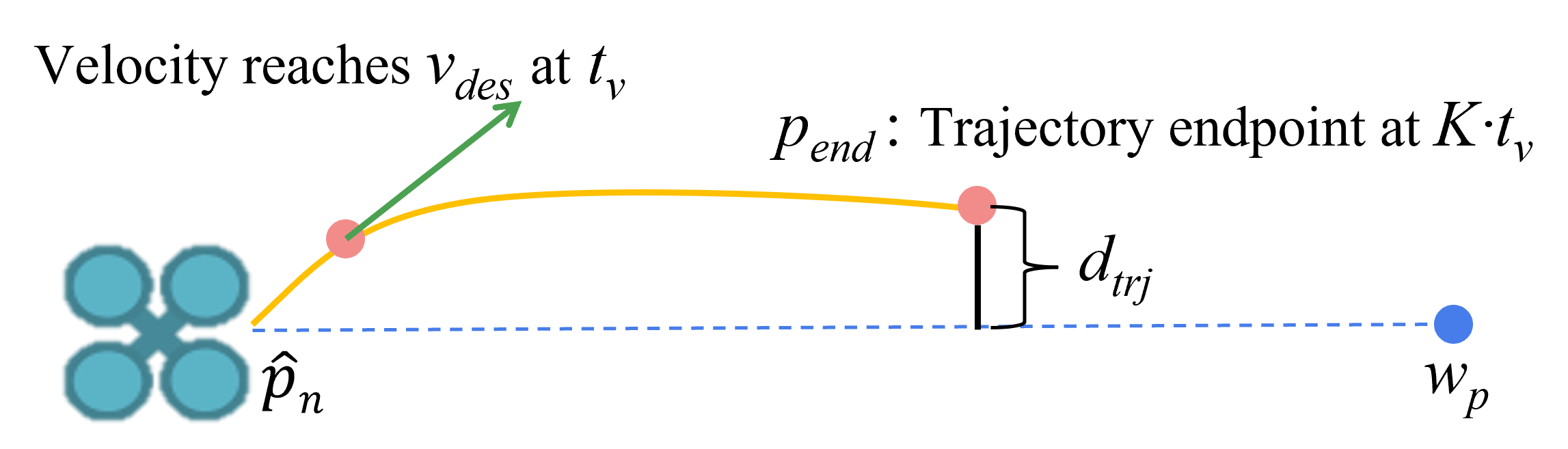}
      \caption{The proposed motion planning method. The objective function is designed to minimize the time cost to reach the desired velocity and the distance from trajectory endpoint $p_{end}$ to the path line. The solid yellow line represents the predicted trajectory.}
      \label{fig_motionplan}
      \vspace{-0.0cm}
   \end{figure}

After the desired velocity $v_{des}$ is obtained, it appears as the constraint in the motion planning and will be reached in a short time. The waypoint constraints $w_{p}$ is also considered to follow the path, as shown in Figure \ref{fig_motionplan}.

The optimization problem to obtain motion primitives is constructed as
\begin{gather*}
\min _{J_{n}, t_{v}} \quad  \eta_{1} t_{v}+\eta_{2} d_{trj}\\
\text{s.t.}\quad a_{n+1} = a_{n} + J_{n} t_{v},\\
v_{des}=v_{n}+a_{n} t_{v} + \frac{1}{2} J_{n} t_{v}^{2},\\
d_{trj} = \frac{\|\overrightarrow{\hat{p}_{n}p_{end}} \times \overrightarrow{w_{p}p_{end}}\|_{2}}{\hat{p}_{n}w_{p}}, \\
p_{end}=\hat{p}_{n}+v_{n}K t_{v}+\frac{1}{2} a_{n} (Kt_{v})^{2} + \frac{1}{6} J_{n} (Kt_{v})^{3},\\
0< \mathrm{t}_{v},\ \left\|a_{n+1}\right\|_{2} \leq a_{\max },\ \|J_{n} \|_{2} < J_{max}, \tag{$20$} \\
\end{gather*}
where the jerk of the vehicle $J_{n}$ is the variable to be optimized. $t_{v}$ is the time required to reach $v_{des}$, which is the variable and the optimization object at the same time. $a_{n+1}$ and $p_{end}$ are calculated by the kinematic formula. $a_{max}$ and $J_{max}$ are the kinodynamic constraints of acceleration and jerk of the vehicle respectively. The velocity constraint $v_{max}$ is satisfied in the equality constraint with $v_{des}$. $\eta_{1},\ \eta_{2}$ are coefficients. The default values are shown in Table \ref{table_2}. After the desired trajectory piece is solved, a default cascade PID controller of PX4 is utilized to track this trajectory in position, velocity, and acceleration.

\section{Experimental implementation and results}
\subsection{Point cloud filters}
\begin{figure}[thpb]
      \centering
      \includegraphics[width=0.99\linewidth]{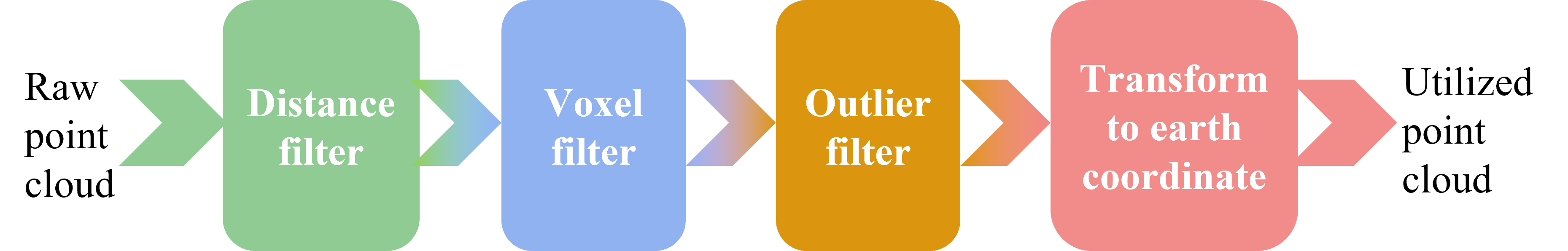}
      \caption{The filtering process for the raw point cloud.}
      \label{fig3}
      \vspace{-0.0cm}
  \end{figure}

For the dynamic environment perception, filtering the raw point cloud is necessary, because the obstacle state estimation is sensitive to the noise. The noise should be eliminated strictly, even losing a few true object points is acceptable. The filter has the same structure as our former work \cite{chen2020computationally}, as shown in Figure \ref{fig3}, but the parameters are different. Distance filter removes the points too far ($\geq 8\ m$) from the camera, voxel filter keeps only one point in one fixed-size ($0.1\ m$) voxel, outlier filter removes the point that does not have enough neighbors ($\leq$ 13) in a certain radius ($0.25\ m$). Based on such configuration, the density threshold for DBSCAN is at least 18 points in the radius of $0.3\ m$. These metrics are tuned manually during extensive tests on the hardware platform introduced in the next subsection, to balance the point cloud quality and the depth detect distance. They are proved satisfactory for obstacle position estimation. The point cloud filtering also reduces the message size by one to two orders of magnitude, so the computation efficiency is much improved, while the reliability of the collision check is not affected.

However, when the drone performs an aggressive maneuver, the pose estimation of the camera (including the ego-motion compensation) is not accurate enough for dynamic obstacle perception. To solve this problem, we propose a practical and effective measurement: The filtered point cloud is accepted only when the angular velocity of the three Euler angles of the drone is within the limit $\omega_{max}$. 

\vspace{-0.0cm}
\subsection{Experimental Configuration}

The detection and avoidance of obstacles are tested and verified in the Robot Operation System (ROS)/Gazebo simulation environment first and then in the hardware experiment. The drone model used in the simulation is 3DR IRIS, and the underlying flight controller is the PX4 1.10.1 firmware version. The depth camera model is Intel Realsense D435i with resolution 424*240 (30 fps). For hardware experiments, we use a self-assembled quadrotor with a QAV 250 frame and an UP board with an Intel Atom x5 Z8350 1.44 GHz processor, other configuration keeps unchanged. Table \ref{table_2} shows the parameter settings for the tests. The supplementary video for the tests has been uploaded online\footnote{https://youtu.be/1g9vHfoycs0}.

\begin{table}[htbp] 
\caption{Parameters for the tests}
\label{table_2}
\begin{center}
\begin{tabular}{cccc}
\toprule
Parameter& Value &Parameter & Value\\
\midrule
$S_{c}$ & 3 & $t_{ct}$ &0.01 s \\
$a_{max}$ & 6 m/s$^2$  & $\omega_{max}$ & 1.5 rad/s\\
$v_{dy}$ & 0.3 m/s & $d_{t}$ & 0.2 s \\

$t_{kf}$ & 0.7 s & $d_{m}$ & 0.9 m \\
$N_{c}$ & 12 &$\alpha$ & 0.5\\
$\eta_1$ & 10 & $\eta_2$ & 6 \\
$ K$ & 3 & $\epsilon$& 0.05 m \\
 \bottomrule
\end{tabular}
\end{center}
\end{table}

\subsection{Simulation Test}

\subsubsection{Dynamic perception module test}
First, the accuracy and stability of the estimation method for the obstacle position and velocity are verified. 

\begin{figure}[h]
\subfigure[]{
\includegraphics[width=0.35\textwidth]{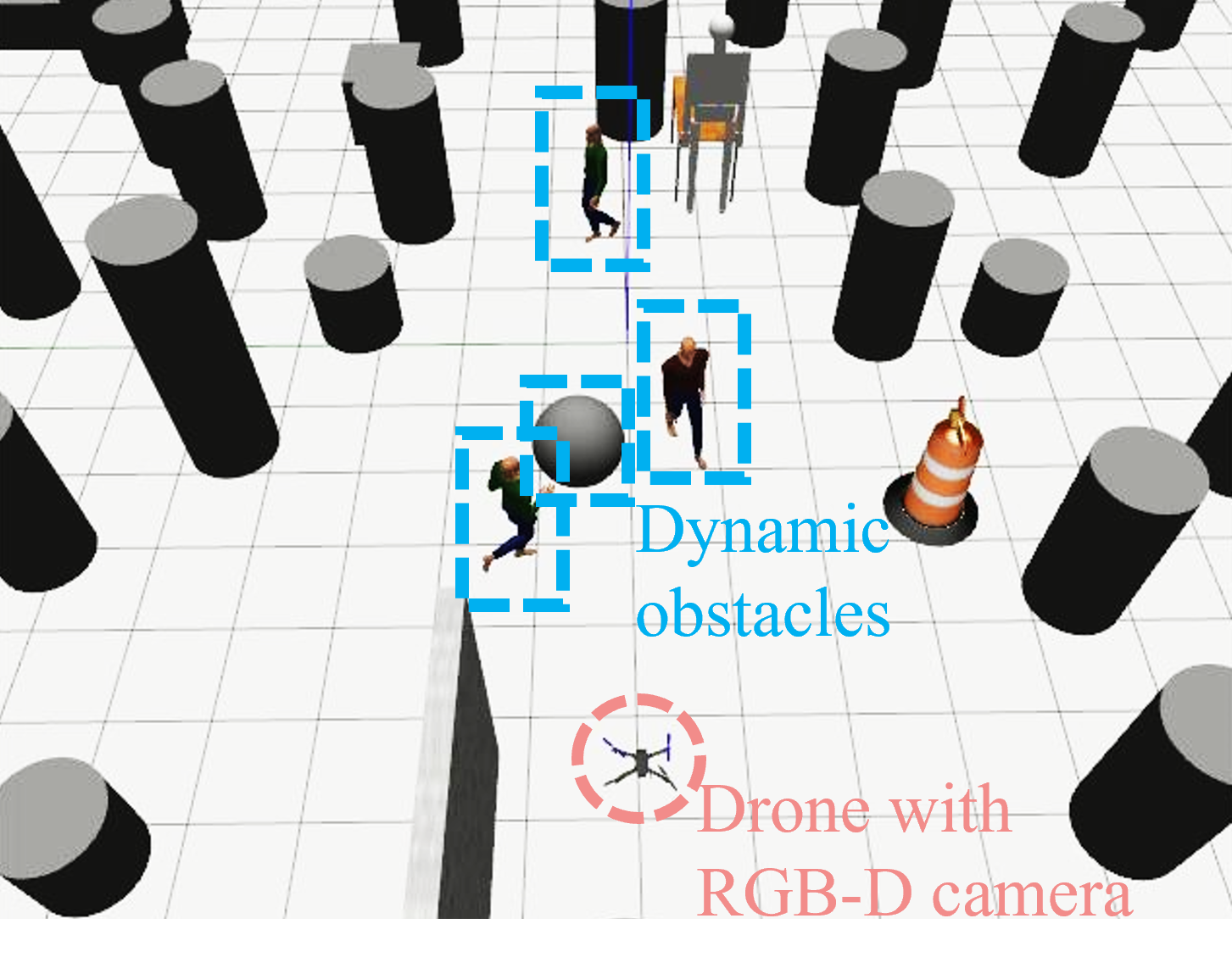}}
\hfill
\centering
\subfigure[]{
\includegraphics[width=0.99\linewidth]{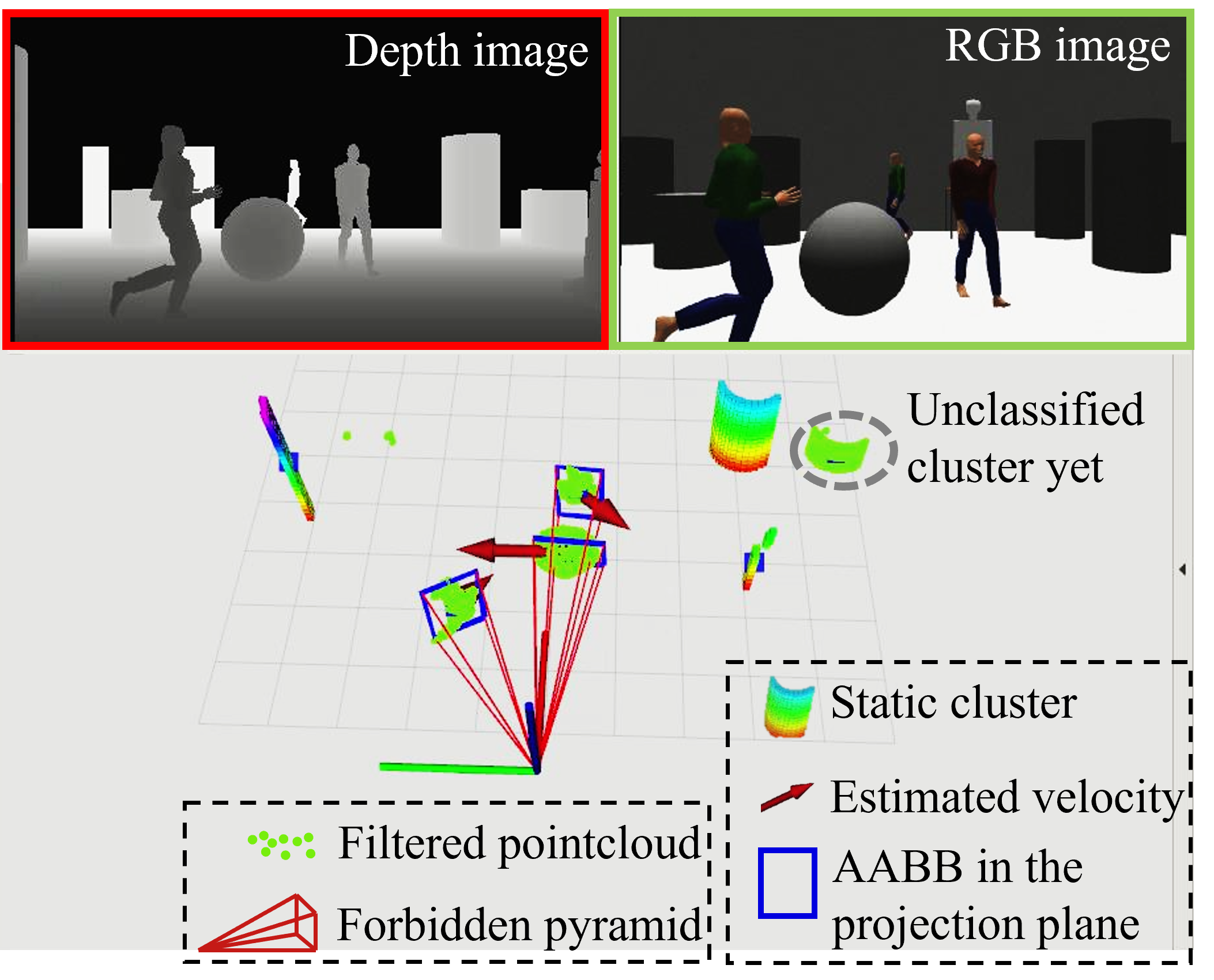}}
\caption{(a): The simulation environment for the moving obstacles' position and velocity estimation test. (b): The visualized estimation results in RVIZ, corresponding to (a). Only the forbidden pyramids for dynamic clusters are visualized. The pedestrians always face their moving direction. It can be seen that the obstacles are correctly tracked even they are very close.}
\label{fig9}
\end{figure}

\begin{figure}[h]
\centering
\includegraphics[width=0.45\textwidth]{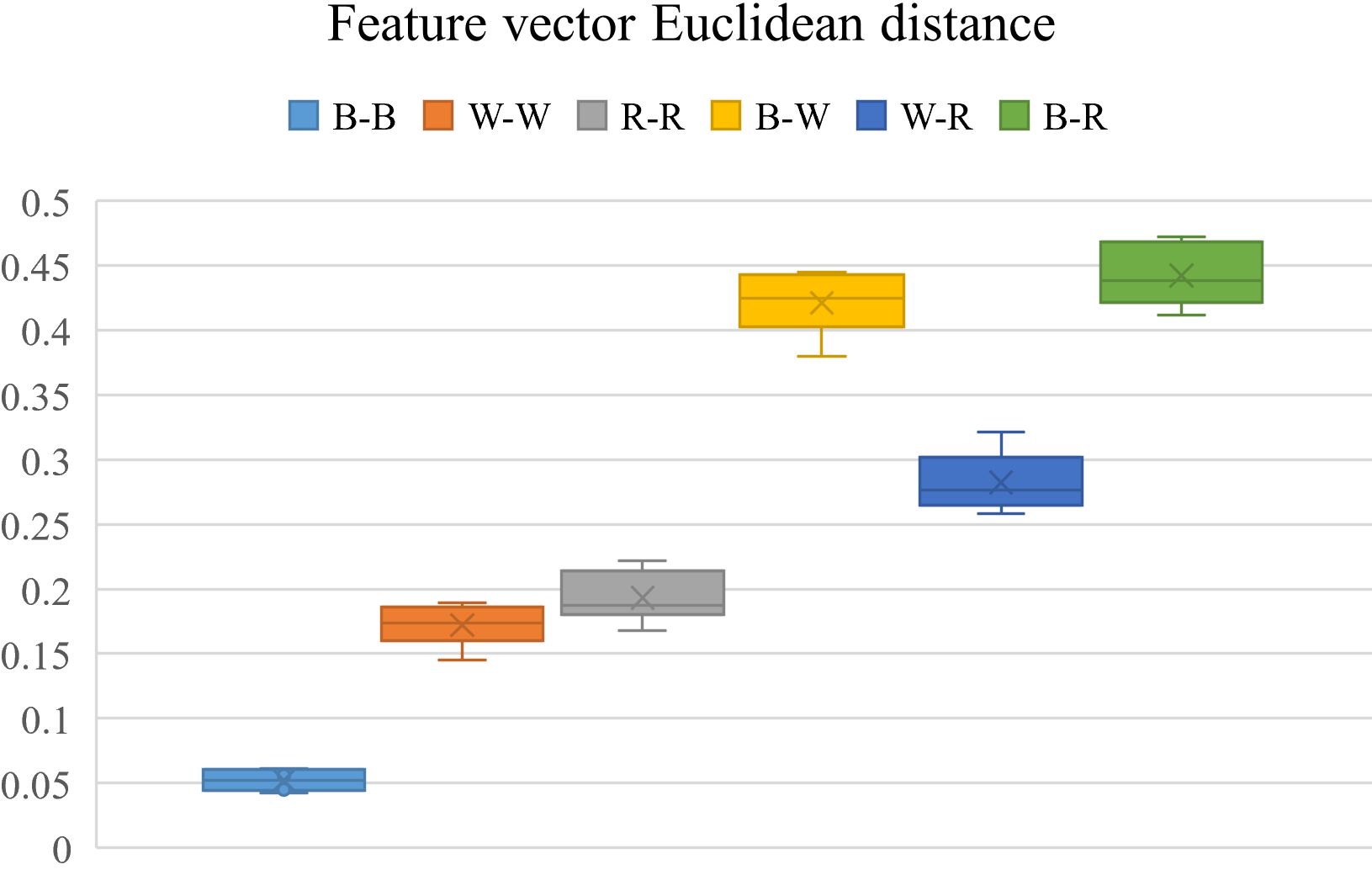}
\caption{The box chart of the Euclidean distance of the feature vector $fte()$ between obstacles from $OB_1$ and $OB_2$. B, W, and R represent the moving \textbf{B}all, \textbf{W}alking and \textbf{R}unning person in Figure \ref{fig9} respectively. The distance of the same obstacle is obviously lower than that of different obstacles, so the obstacles are matched correctly.}
\label{fig-boxchart}
\vspace{-0.0cm}
\end{figure}

\begin{figure}[h]
\includegraphics[width=0.46\textwidth, height = 5.5cm]{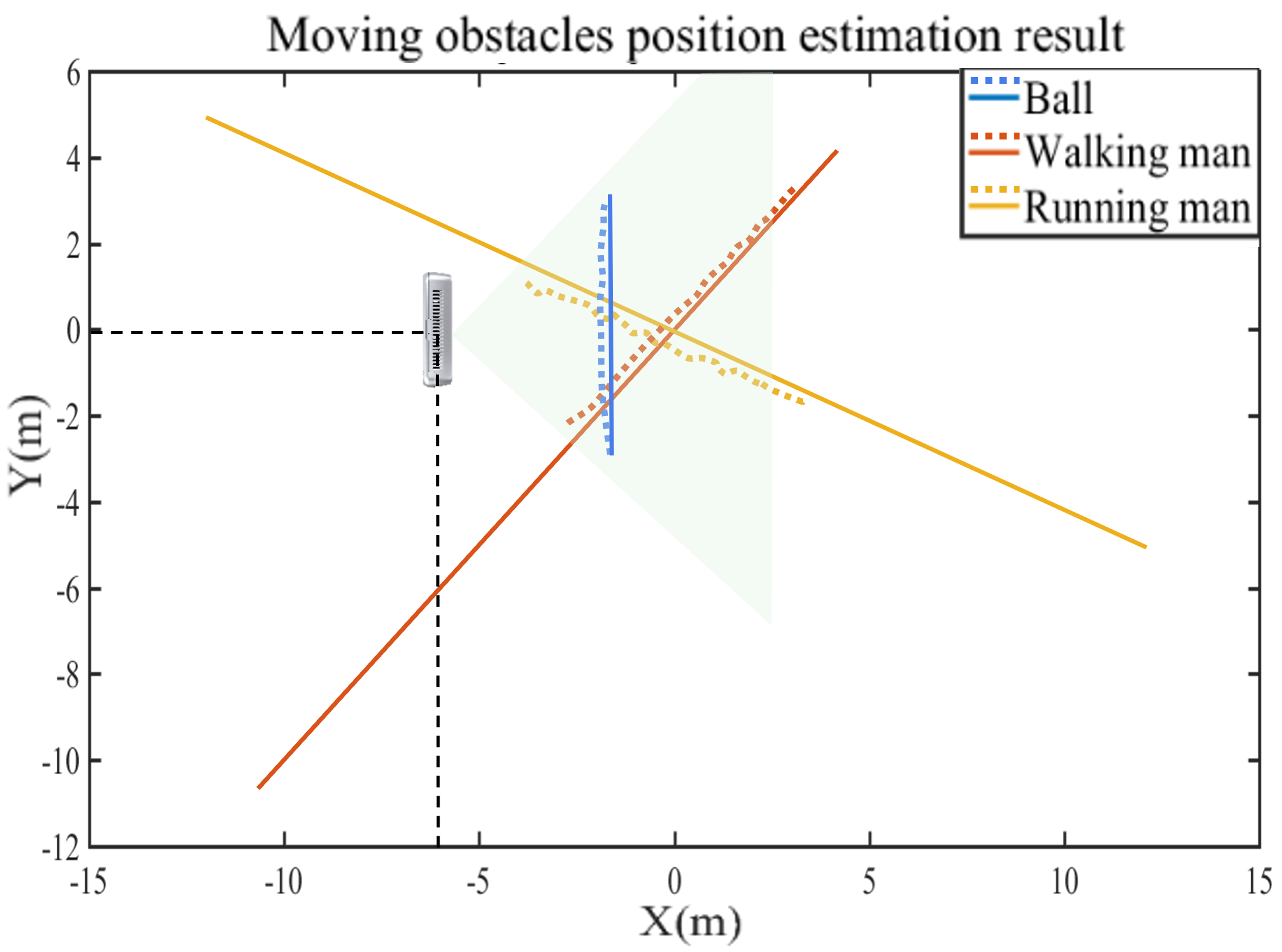}
\caption{The estimation results of the moving obstacles position. The FOV of the camera is represented with light green area. The dotted line is the estimated result, while the solid line is the ground truth.}
\label{fig10}
\vspace{-0.0cm}
\end{figure}

In the simulation world depicted in Figure \ref{fig9}(a), there is one moving ball, two moving human models, and some static objects. The moving obstacles are reciprocating on different straight trajectories. The camera is fixed on the head of the drone, facing forward straightly. The drone is hovering around the point $(-6,0,1.2)$. Figure \ref{fig9}(b) depicts the visualized estimation results in Rviz. The Euclidean distance of the feature vectors utilized for obstacle tracking is illustrated in Figure \ref{fig-boxchart}. 
The estimation numeral results are shown in Figure \ref{fig10}, and they are compared with the ground truth. The dynamic perception performance is also compared with SOTA works in Table \ref{table_state_estimate}, where the metrics Multiple Object Tracking Precision (MOTP) and Multiple Object Tracking Accuracy (MOTA) are adopted as defined in the work of Bernardin \cite{bernardin2006multiple}. MOTP is the average position estimation error in this test. Only walking or running pedestrians are tested in Table \ref{table_state_estimate}. The second line marked with * is for our method without using the track point to correct the velocity estimation, and the third line marked with \# is for our method without the neighbor frame overlapping.

The estimation test results demonstrate that our estimation algorithm is practical for dynamic obstacle avoidance. In addition, our method efficiently improves the estimation accuracy and robustness in the clustered environment. For our MOTA, it is composed of a false negatives rate $f_{n} = 6.7\%$ (covering non-detected dynamic objects and dynamic objects erroneously classified as static or uncertain), a false positives rate $f_{p} = 6.9\%$ (static objects misclassified as dynamic), and a mismatch rate $f_{m} = 2.1\%$. 

\vspace{-0.0cm}

\begin{figure}[h]
\includegraphics[width=0.99\linewidth]{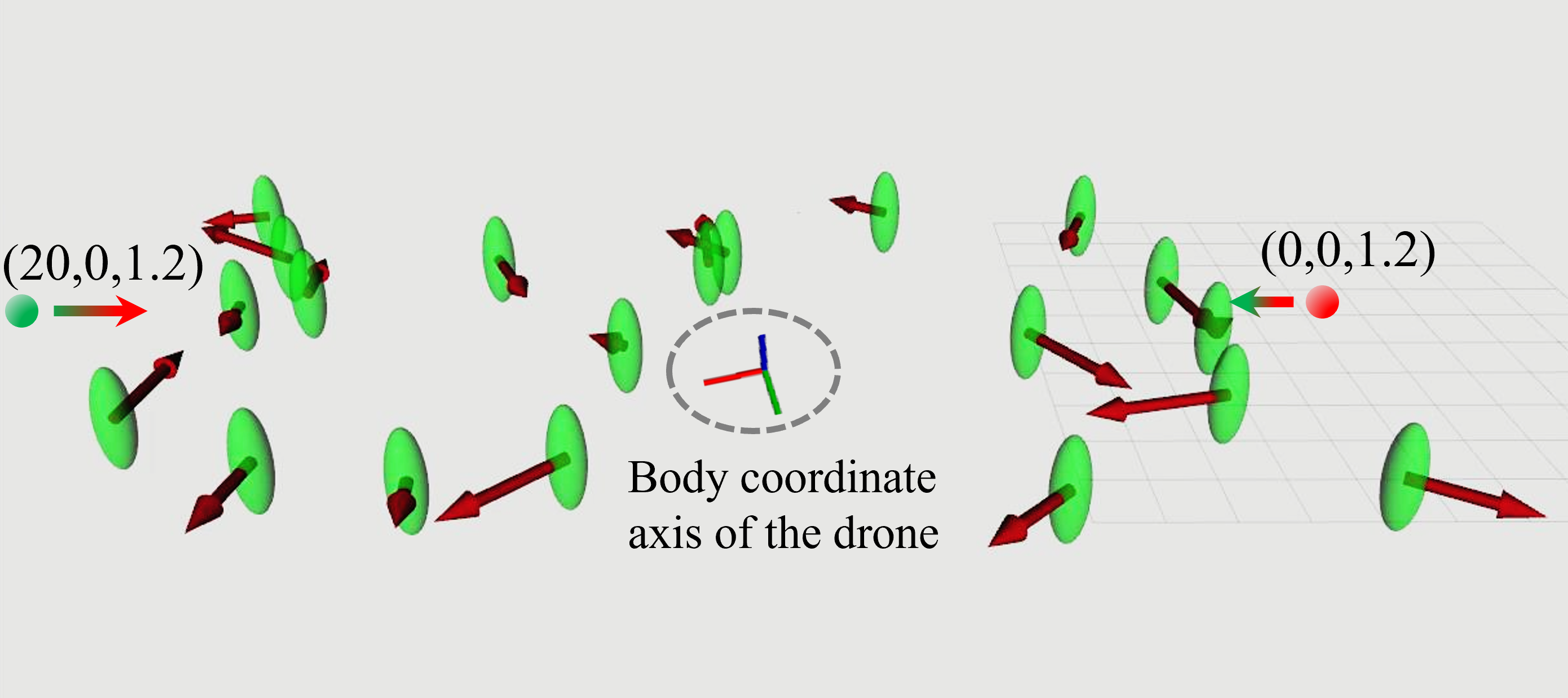}
\caption{The simulated test environment for the motion planning module. The drone flies between the two points for the assigned times.}
\label{motion_plan_test}
\end{figure}

\begin{table}[h]
\caption{Obstacle State Estimation Comparison}
\label{table_state_estimate}
\begin{center}
\begin{tabular}{cccc}
\toprule
Method & $error_{vel}(m/s)$ & MOTA ($\%$) & MOTP\\
\midrule
\textbf{Ours} & \textbf{0.21} & \textbf{84.3} & \textbf{0.15} \\

Ours* & 0.29 & 83.9 & 0.16  \\

Ours\# & 0.25 & 83.6 & 0.18  \\

\cite{wang2021autonomous} & 0.37 & 76.4 & 0.28  \\

\cite{lin2020robust} & 0.41 & 70.1 & 0.30 \\

\bottomrule
\end{tabular}
\end{center}
\end{table}

\begin{table}[h]
\caption{Dynamic Planning Comparison}
\label{table_motion}
\begin{center}
\begin{tabular}{p{1cm}<{\centering}p{1.6cm}<{\centering}p{1.5cm}<{\centering}p{1cm}<{\centering}p{1cm}<{\centering}}
\toprule
Method & $a_{mean}(m/s^{2})$ & $v_{mean}(m/s)$ & $l_{traj}(m)$ & $t_{opt}(ms)$\\
\midrule
\textbf{Ours} & \textbf{2.96} & 2.24 & 25.21 & \textbf{3.15}\\

\cite{wang2021autonomous} & 3.43 & 2.37 & 23.65 & 8.61  \\ 

\cite{zhu2019chance} & 3.18 & 2.33 & 22.96 & 31.23\\ \bottomrule
\end{tabular}
\end{center}
\end{table}

\begin{figure}[thpb]
\centering
\subfigure[]{
\includegraphics[width=0.22\textwidth,height = 3cm]{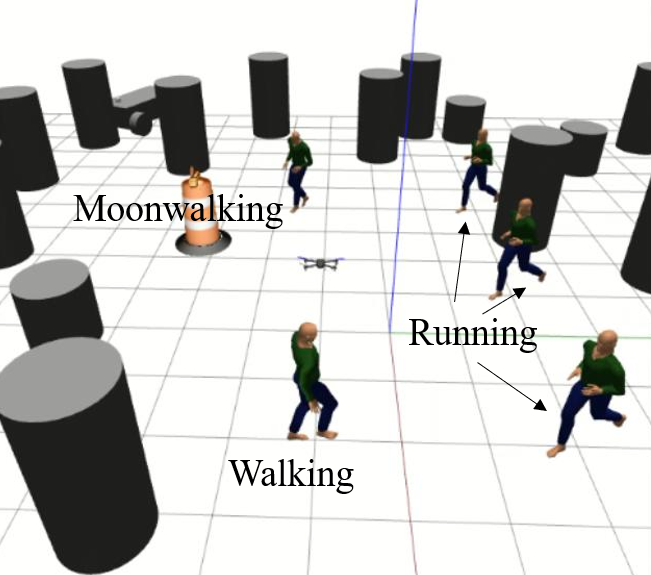}}
\hfill
\centering
\subfigure[]{
\includegraphics[width=0.24\textwidth,height = 3cm]{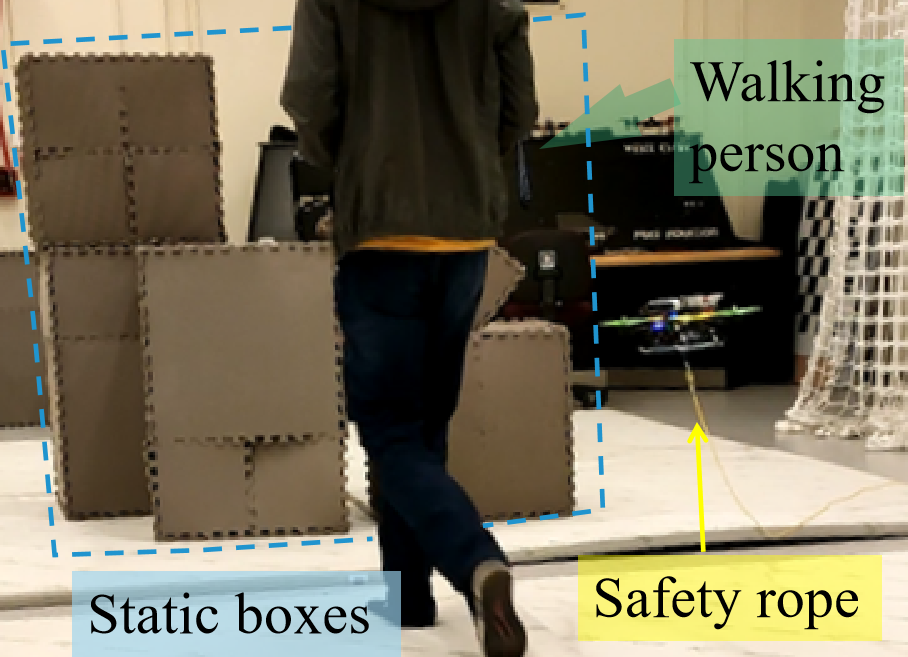}}
\hfill
\centering
\subfigure[]{
\includegraphics[width=0.46\textwidth,height = 4.7cm]{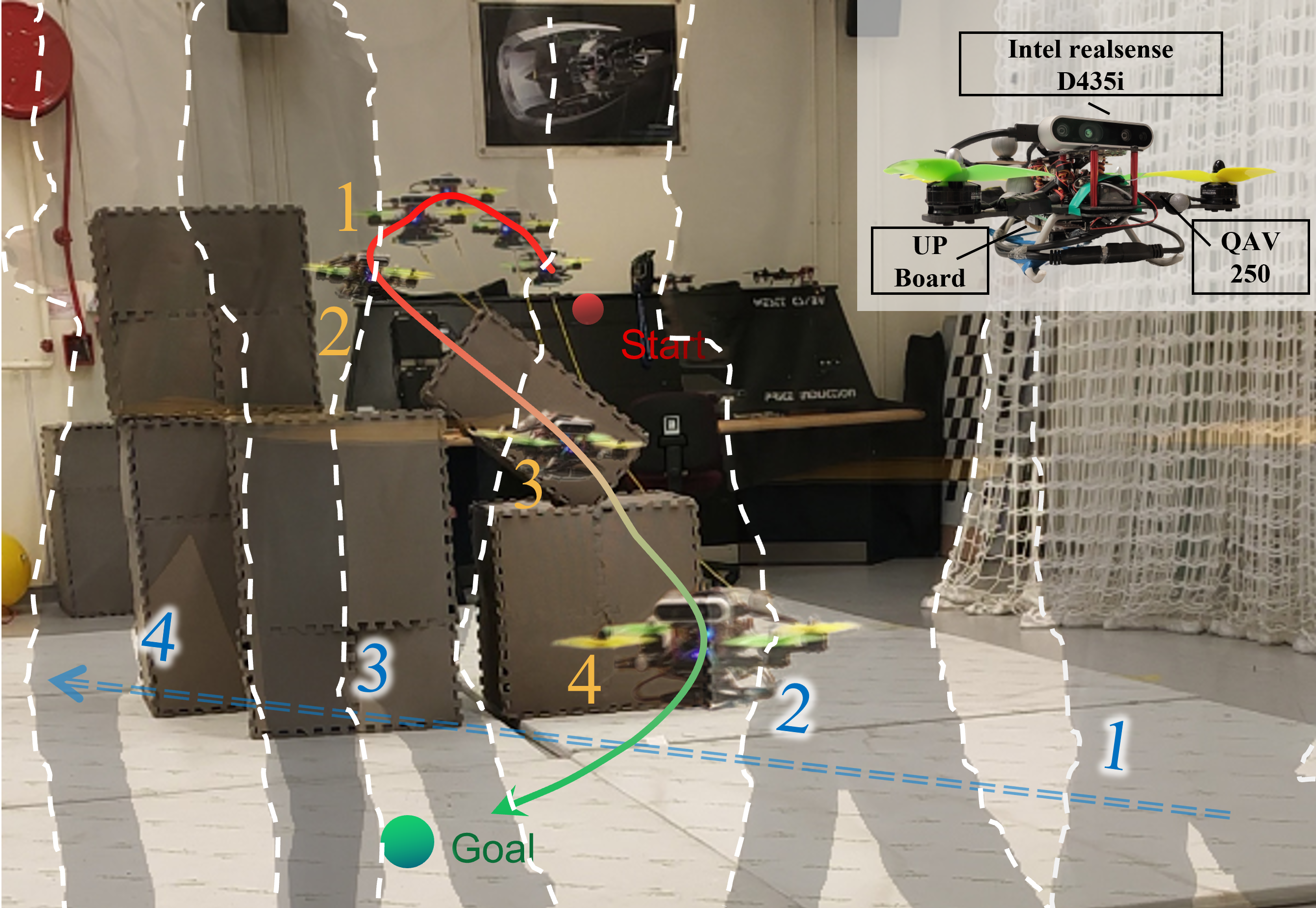}}
\vspace{-0.0cm}
\caption{(a): The dynamic simulation test environment. (b): The dynamic hardware test environment. (c): The composed picture of the hardware flight test. Numbers in (c) mark the same frame for the drone and person.}
\label{fig11}
\vspace{-0.0cm}
\end{figure}

 \subsubsection{Motion planning module test}
 In addition, we compare the motion planning method with \cite{wang2021autonomous} and \cite{zhu2019chance} in Table \ref{table_motion}. Because locations and velocities of all obstacles are known and not classified as dynamic or static in \cite{zhu2019chance}, the comparison for the planning module in an environment with only dynamic obstacles. The obstacles are ellipsoids with human-like size (0.5×0.5×1.8m) and move with constant velocities, as shown in Figure \ref{motion_plan_test}. For each planner, the drone flies between two points $(0,0,1.2)$ and $(20,0,1.2)$ back and forth for 10 times, 20 obstacles with velocities at 1-3 $m/s$ cross this straight path disorderly. The camera FOV is also considered, and simulated to be $85.2^{\circ} \times 58^{\circ}$ with the maximal sensing depth $8\ m$. From Table \ref{table_motion} we conclude that the average acceleration cost of our motion planning method is smaller because our velocity planning method considers the minimal acceleration cost in all the sample velocities for the current time. Also, the computing time is much shorter thanks to the simple but efficient object function and constraints, which shows the potential to avoid faster obstacles.

\subsubsection{System test}
To test the whole framework for navigation tasks, we utilize the HAS method \cite{chen2020computationally} as the path planning algorithm at the front end to generate the waypoint $w_{p}$ for (20). Figure \ref{fig11}(a) is the overview of the flight simulation world. In Figure \ref{fig1}, the necessity for estimating the obstacle's velocity is illustrated: To avoid the moving man which is at a similar speed with the drone, the aircraft choose to fly in the ``opposite'' direction with the man so the threat is removed easily. If only the static HAS method \cite{chen2020computationally} is utilized in the same situation, the drone decelerates and fly alongside the man (red line), which is very inefficient and dangerous. 

\subsection{Hardware Flight Test}
\begin{figure}[thpb]
      \centering
      \subfigure[]{
      \includegraphics[width=0.99\linewidth,height =2.7cm]{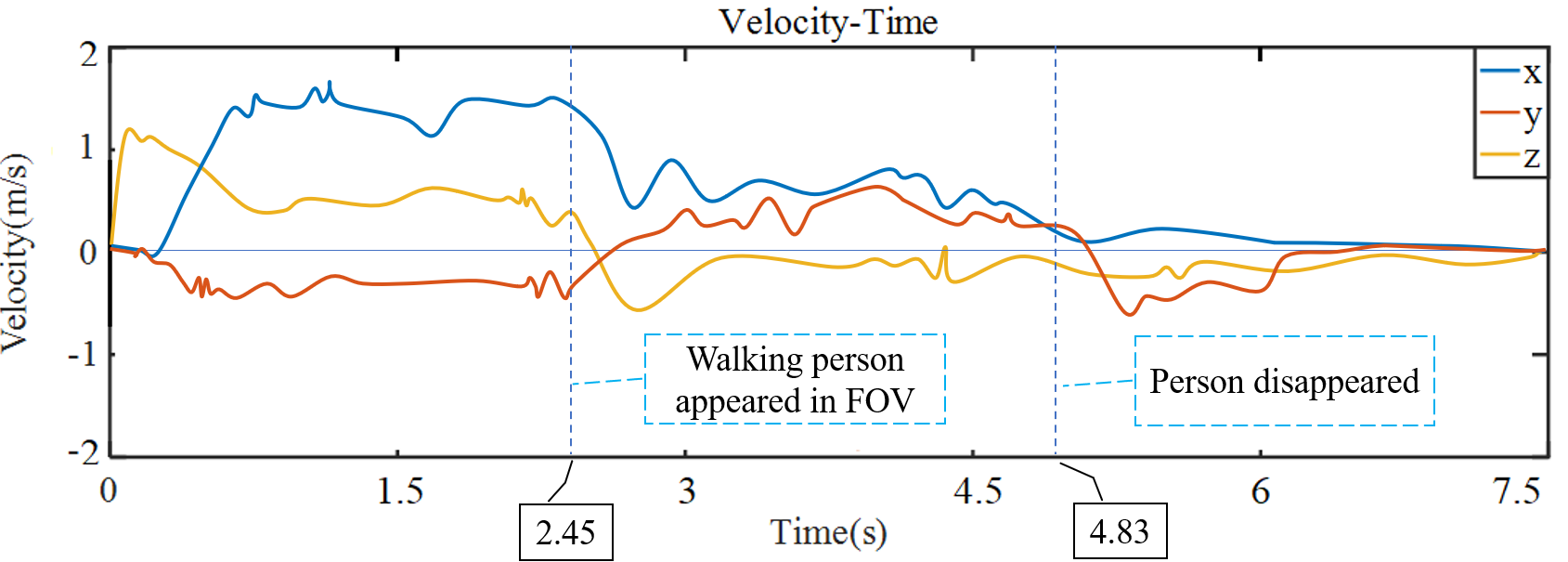}}
    \centering
      \subfigure[]{
           \includegraphics[width=0.99\linewidth,height =2.5cm]{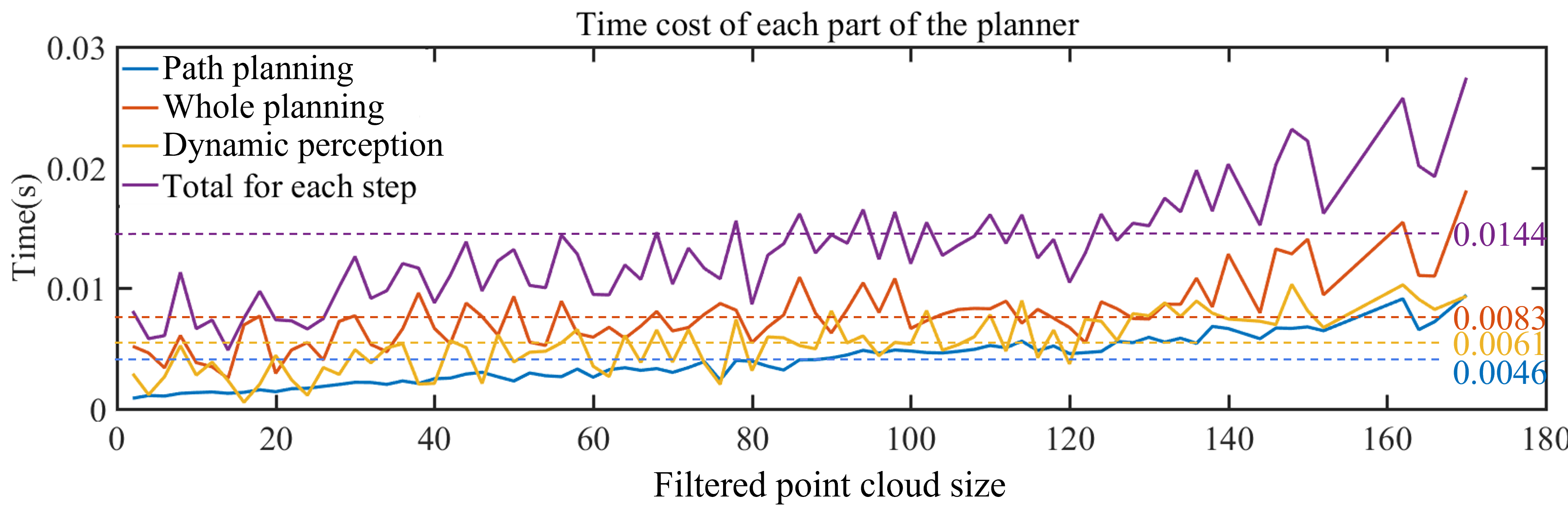}}
     \caption{(a): The velocity of the drone in earth coordinate $E\!-\!X\!Y\!Z$. (b): The time cost for different modules under different filtered point cloud size. The average value is marked in the corresponding color.}
      \label{fig12}
   \end{figure}

 We set up a hardware test environment as Figure \ref{fig11}(b), the drone takes off behind the boxes and then a person enters the FOV of the camera and walks straight after the drone passes static boxes to test the effectiveness of our method. In Figure \ref{fig11}(c), the camera is fixed and takes photos every 0.7 s during the flight. Seven photos are composed together. In the 4th frame the walking person appeared, the solid line shows the trajectory of the drone while the blue dashed line is for the person. It can be concluded that the reaction of the drone is similar to the simulation above. The data for this hardware test is shown in Figure \ref{fig12}, the velocity curve indicates that the drone reacts very fast once the moving obstacle appears. In Figure \ref{fig12}(b), the blue line is only for the path planning algorithm (HAS), and the red line represents the whole planning with \textbf{Algorithm \ref{alg5}}. The blue line has a strong positive linear correlation with time because the number of the input points of the collision check procedure determines the distance calculation times. The red and yellow line show the irregularity, which is because the moving obstacle brings external computation burden to \textbf{Algorithm \ref{alg5}}, and the number of moving obstacles has no relation with the point cloud size. The single-step time cost of our proposed method (exclude the path planning) is even smaller than 0.01 s, indicating the fast-reacting ability towards moving obstacles.

At last, we compare our work with SOTA works on the system level in Table \ref{table_3}. Since most related works differ significantly from ours in terms of application background and test platform, for numeral indicators we only compare the total time cost for reference. The abbreviations stand for: obs (obstacle), cam (camera), UUV (underwater unmanned vehicle), N/A (not applicable). ``N/A'' in the sensor type column refers to the work that gets obstacle information from an external source and does not include environment perception. Most works have severe restrictions on the obstacle type or incompleteness in environment perception, and the computing time cost is not satisfactory for real-time applications. Our work has a great advantage in generality and system completeness, the time cost is also the shortest unless the event camera is considered.

\begin{table}[h]
\caption{System comparison between different works}

\label{table_3}
\begin{center}
\begin{tabular}{p{0.5cm}<{\centering}p{2.4cm}<{\centering}p{0.9cm}<{\centering}p{2.0cm}<{\centering}p{1.6cm}<{\centering}}
\toprule
Work & Sensor type & Vehicle & Obs limits & Time cost (s) \\
\midrule
\cite{zhang2017dynamic} & Sonar & UUV & dynamic obs & 1-2\\

\cite{malone2017hybrid} & N/A & Robots & dynamic obs & 0.045-0.13\\

\cite{cherubini2014autonomous} & Cam \& Lidar & Car & \textbf{N/A} & 0.1 \\

\cite{nageli2017real}& N/A & UAV & human & 0.2-0.3 \\

\cite{falanga2020dynamic} & Event cam & UAV& dynamic obs & \textbf{0.0035} 
\\

\cite{lin2020robust} & N/A & UAV & \textbf{N/A} & 0.07 \\

\textbf{Ours}&RGB-D cam & UAV & \textbf{N/A} & \textbf{0.015}\\

\bottomrule
\end{tabular}
\end{center}
\end{table}

\section{Conclusion and future work}

In this paper, we present a computationally efficient algorithm framework for both static and dynamic obstacle avoidance for UAVs based only on point cloud. The test results indicate our work is feasible and shows great promise in practical applications.

However, when the speed or angular velocity of the drone is high, and also because of the narrow FOV of a single camera, the dynamic perception becomes significantly unreliable. In future research, we intend to improve the robustness of our method in aggressive flights and test it with different sensors such as lidar.

\bibliographystyle{elsarticle-num}
\bibliography{cas-dc-template.bbl}




\end{document}